\documentclass{article}

\PassOptionsToPackage{}{natbib}

\usepackage[preprint]{neurips_2024}

\usepackage[table]{xcolor} 
\usepackage{colortbl}
\usepackage{graphicx}   
\usepackage{subcaption} 
\usepackage[utf8]{inputenc} 
\usepackage[T1]{fontenc}    
\usepackage{hyperref}       
\usepackage{url}            
\usepackage{booktabs}       
\usepackage{amsfonts}       
\usepackage{amsmath}
\usepackage{nicefrac}       
\usepackage{microtype}      
\usepackage{xcolor}         
\usepackage{enumitem} 
\usepackage{multirow}
\usepackage{adjustbox}
\usepackage{caption}
\usepackage{bm}
\usepackage[detect-all]{siunitx}
\usepackage[linesnumbered,ruled,vlined]{algorithm2e}
\usepackage[normalem]{ulem}
\usepackage[most]{tcolorbox}
\usepackage[page,header]{appendix}
\usepackage{titletoc}

\newtcolorbox{promptbox}{
    colback=blue!5!white,    
    colframe=blue!50!black,  
    fonttitle=\bfseries,
    title=Prompt,            
    sharp corners,           
    boxrule=0.8pt,           
    left=6pt,                
    right=6pt,
    top=4pt,
    bottom=4pt,
    breakable                
}
\DeclareRobustCommand{\vect}[1]{\bm{#1}}
\definecolor{saepink}{RGB}{255, 213, 214}
\definecolor{induction}{RGB}{230, 240, 255} 
\definecolor{specific}{RGB}{240, 230, 255}  
\definecolor{previous}{RGB}{255, 240, 230}  
\definecolor{acronym}{RGB}{240, 255, 240}   
\definecolor{foreign}{RGB}{255, 250, 230}   
\definecolor{sink}{RGB}{255, 230, 240}      
\definecolor{other}{RGB}{230, 255, 250}     

\title{Towards Understanding the Nature of Attention with Low-Rank Sparse Decomposition}

\author{
  Zhengfu He\textsuperscript{1,2}\thanks{Equal Contribution.}\quad
  Junxuan Wang\textsuperscript{1,2}\footnotemark[1] \quad
  Rui Lin\textsuperscript{1,2} \quad
  Xuyang Ge\textsuperscript{1,2} \quad \\
  \textbf{Wentao Shu\textsuperscript{1,2}} \quad 
  \textbf{Qiong Tang\textsuperscript{2}}\quad
  \textbf{Junping Zhang\textsuperscript{2}}\quad
  \textbf{Xipeng Qiu\textsuperscript{1,2}}\thanks{Corresponding Author.} \quad \\\\
  \textsuperscript{1}Shanghai Innovation Institute \\
  \textsuperscript{2}OpenMOSS Team, School of Computer Science, Fudan University \\\\
  \texttt{zfhe19@fudan.edu.cn}
}

\begin{document}

\maketitle

\begin{figure}[h]
  \centering
  \includegraphics[width=\textwidth]{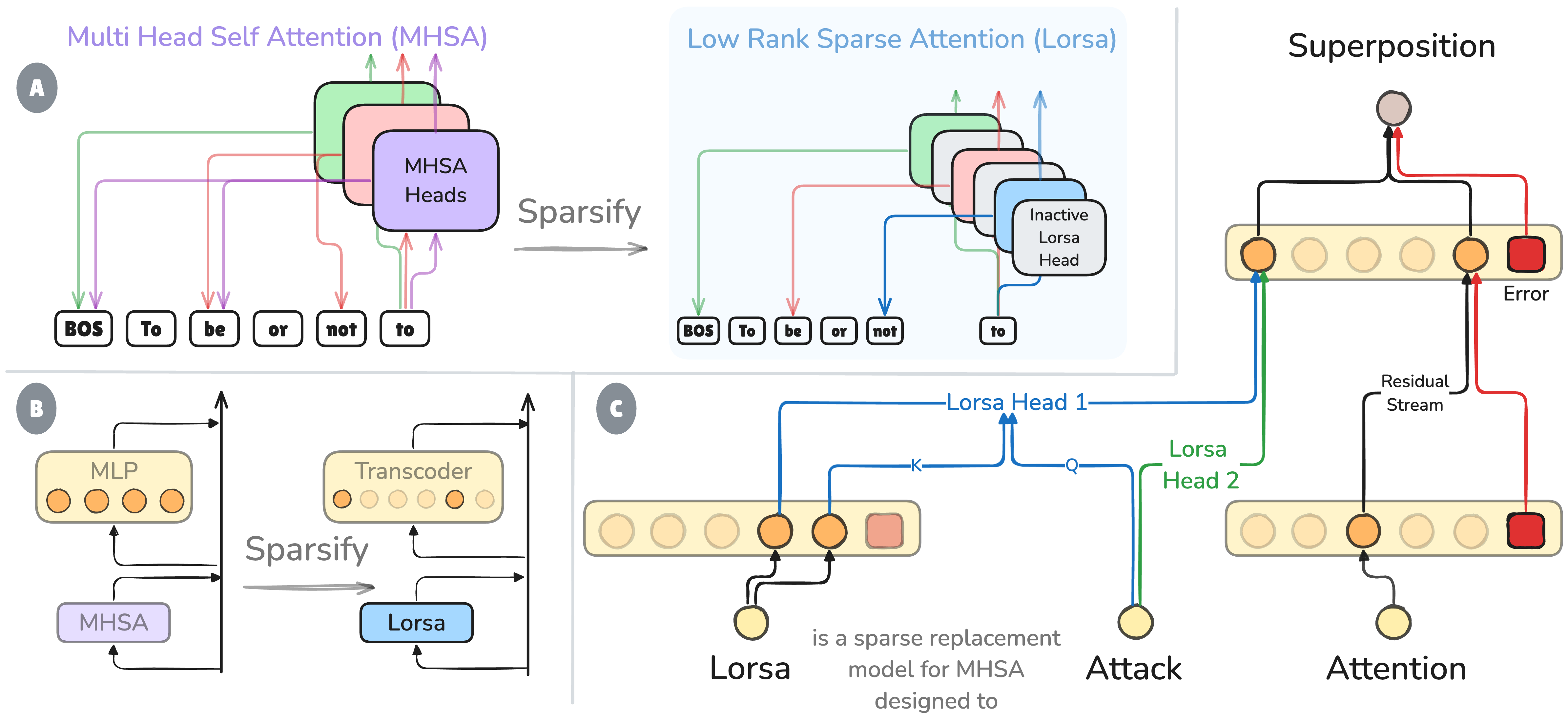}
  \caption{(\textbf{A}) \textcolor{cyan}{Lo}w-\textcolor{cyan}{R}ank \textcolor{cyan}{S}parse \textcolor{cyan}{A}ttention (Lorsa) comprises thousands of sparsely activated attention heads with 1D
  outputs, designed to extract interpretable attention units from the original Multi Head Self Attention (MHSA). (\textbf{B}) Lorsa
  serves as a replacement model for Transformer attention, substituting sparse interpretable components for attention modules.
  (\textbf{C}) Each Lorsa head explains an atomic feature-feature interaction across token
  positions, which was originally a part of an MHSA head or spread across multiple heads, i.e. put in attention superposition.}
  \label{fig:head_img}
\end{figure}

\begin{abstract}
We propose \textcolor{cyan}{Lo}w-\textcolor{cyan}{R}ank \textcolor{cyan}{S}parse \textcolor{cyan}{A}ttention (Lorsa), a sparse replacement model
of Transformer attention layers to disentangle original Multi Head Self Attention (MHSA) into individually comprehensible components.
Lorsa is designed to address the challenge of \textit{attention superposition} to understand attention-mediated interaction between
features in different token positions. We show that Lorsa heads
find cleaner and finer-grained versions of previously discovered MHSA behaviors like induction heads, successor heads and attention sink
behavior (i.e., heavily attending to the first token). Lorsa and Sparse Autoencoder (SAE) are both sparse dictionary learning methods
applied to different Transformer components, and lead to consistent findings in many ways. For instance, we discover a comprehensive family
of arithmetic-specific Lorsa heads, each corresponding to an atomic operation in Llama-3.1-8B. Automated interpretability analysis indicates
that Lorsa achieves parity with SAE in interpretability while Lorsa exhibits superior circuit discovery properties, especially for features
computed collectively by multiple MHSA heads. We also conduct extensive experiments on architectural design ablation, Lorsa scaling law and
error analysis.
\vspace{-0.3em}
\begin{center}
  \texttt{Code:} \url{https://github.com/OpenMOSS/Lorsa}\\
  \texttt{Lorsa Weights: \url{https://huggingface.co/fnlp/Lorsa}} 
\end{center}
  
\end{abstract}

      

\section{Introduction}
\label{sec:intro}


When examining the function of individual attention heads in a Transformer model, one might identify some of these
heads implementing a specific behavior. A canonical example is induction heads which predicts 'Potter' following the token
'Harry' when 'Harry Potter' is present in the context~\citep{olsson2022inductionheads}. Ablating these heads substantially
prevents the model from correctly performing corresponding tasks, which indicates causal relation of these heads and the
model's macroscopic behaviors. These interpretable attention units constitute the basic building blocks of the model's
inter-token information mixing algorithm.

Not all attention heads, however, exhibit clear functionality. Most heads distribute attention across diverse contexts.
Although some heads exhibit identifiable patterns, there might be inter-head collaboration that explains the whole story.
These challenges in attention head interpretation is analogous to feature superposition in understanding individual neurons, which
suggests the existence of \textbf{attention superposition}~\citep{jermyn2024attentionsuperposition} in Multi Head Self Attention (MHSA),
which we will further discuss in Section~\ref{sec:att_superp}.

Inspired by the recent success of Sparse Autoencoders (SAEs) to extract monosemantic features from Transformers' hidden
space~\citep{templeton2024scaling} or approximate part of the
network's computation as a sparse computation~\citep{templeton2024predictfuture,ge2024hierattr,dunefsky2024transcoder}, we propose
\textcolor{cyan}{Lo}w-\textcolor{cyan}{R}ank \textcolor{cyan}{S}parse \textcolor{cyan}{A}ttention (Lorsa) to disentangle the atomic
attention units from attention superposition (Section~\ref{sec:lorsa}). Lorsa serves as a replacement module of the original MHSA with an overcomplete set of
attention heads featuring a single-dimensional OV circuit~\citep{elhage2021framework} and sparsity constraints.

In Section~\ref{sec:interpret_lorsa_heads}, we introduce our exploration interface following~\citet{bricken2023monosemanticity},
providing multifaceted information on each Lorsa head. We also quantitatively assess Lorsa head interpretability using top activations
and their attribution patterns ($z$ pattern) with automated interpretability~\citep{bills2023autointerp}. The results indicate that Lorsa's
monosemanticity is comparable to SAE features.

Section~\ref{sec:search_specific_lorsa_heads} presents findings with Lorsa on Pythia-160M~\citep{Biderman2023Pythia} and Llama-3.1-8B~\citep{dubey2024llama3}.
For validation, we first identify the Lorsa instantiations of known attention mechanisms: \emph{induction heads}, \emph{name mover heads}~\citep{wang23ioi},
\emph{successor heads}~\citep{gould24successorhead}, and attention sinks~\citep{xiao2024attentionsink}. Furthermore, we characterize
a family of arithmetic-specific Lorsa heads in Llama-3.1-8B. We also
identify a subset of Lorsa heads in Llama-3.1-8B that function as \emph{thematic anchors} by exhibiting long-range, topic-specific attention patterns.

To the best of our knowledge, Lorsa is the first attempt to extract sparse and interpretable attentional computation, yet still has
significant room for improvement in aspects discussed in Section~\ref{sec:discussion}. We hope these discussions and findings will
facilitate future research along this direction.

\textbf{Note on Terminology:} While prior work refers to the atomic computational units we aim to independently understand as
\emph{attentional features}~\citep{jermyn2024attentionsuperposition,ameisen2025circuit}, we adopt \emph{attention units}
to avoid conflating with activation-space features (which denote 1D linear features in representation spaces~\citep{elhage2022tms}).
The term \emph{head} flexibly denotes either MHSA heads or Lorsa heads as context dictates.

\section{Attention Superposition}
\label{sec:att_superp}
Analogous to how post-ReLU neurons in Transformer MLPs learn to represent more features than they have dimensions~\citep{elhage2022tms},
a similar phenomenon may occur in Multi-Head Self Attention (MHSA). We hypothesize MHSA may comprise multiple attention units in
\textbf{attention superposition}, each attending between certain token pairs with interpretable read/write operations on the residual stream.
Under this hypothesis, we would expect (1) an atomic attention unit is spread across multiple MHSA heads. (2) One MHSA head includes
multiple units. We list three points of evidence of attention superposition in Transformer language models.


\textbf{1. A Few Neurons (Heads) Are Polysemantic.}
\citet{gurnee23neurons_haystack} discovered compound word neurons activating
across diverse unrelated n-grams, while~\citet{bricken2023monosemanticity} reported neurons responding to mixed stimuli including academic
citations and Korean text.(\href{https://transformer-circuits.pub/2023/monosemantic-features/vis/a-neurons.html?ordering=index#feature-83}{link}).
Similarly, successor heads~\citep{gould24successorhead} which increment ‘Monday’ into ‘Tuesday’ and ‘1’ into ‘2’ simultaneously exhibit
Acronym behavior, Copying behavior and Greater-than behavior.

\textbf{2. Most Neurons (Heads) Exhibit Uninterpretable Activating (Attention) Patterns.}
Multiple studies report the predominance of MLP neurons lacking clear activation patterns~\citep{arora2018superposition,bricken2023monosemanticity}.
Likewise,~\citet{krzyzanowski2024inspect_every} reports failed interpretation attempts for more than 90\% heads in GPT-2.

\textbf{3. Attention Superposition in the Wild.}
\citet{he2024othello} and~\citet{kissane2024attentionsae} both found attention output SAE features collectively contributed by multiple
attention heads. If we consider SAE features to represent monosemantic directions, such distribution provides evidence for
attention superposition. Furthermore,~\citet{jermyn2024attentionsuperposition} directly demonstrate this phenomenon through
a constructed case where 5 ground-truth attention units are put in superposition over 2 attention heads. We also show that about
25\% of our learned attention units are spread across multiple MHSA heads (Appendix~\ref{appendix:corr_wiz_MHSA:statistics}).

\paragraph{Why Does Attention Superposition Matter?}
Practically, attribution-based circuit tracing~\citep{ge2024hierattr,ameisen2025circuit}
becomes challenging when features are computed collectively: individual QK patterns do not explain the full mechanism and may be
misleading due to interference from other features' computations within the same heads. The structure of attention superposition
may relect intriguing motifs of model biology. For example, what makes some privleged attention units like induction heads mostly
implemented by a single MHSA head~\citep{olsson2022inductionheads} while others are put
in superposition? This parallels privleged bases in MLP neurons~\citep{elhage2023privbasis}.

\section{Low-Rank Sparse Attention}
\label{sec:lorsa}

\subsection{Lorsa Architecture}
\label{sec:lorsa:arch}

\begin{algorithm}[h!]
  \caption{Low-Rank Sparse Attention (\sout{MHSA} \textcolor{cyan}{Lorsa})}
  \label{alg:lorsa}
  \KwIn{
      $\mathbf{X} \in \mathbb{R}^{n \times d}$: Input sequence (n tokens, d dimensions) \\
      $W_q^h, W_k^h \in \mathbb{R}^{d \times d_h}$: Query/Key weights for head $h$ \\
      \sout{$W_v^h \in \mathbb{R}^{d \times d_h}$} $\textcolor{cyan}{w_v^h} \in \mathbb{R}^{d \times \textcolor{cyan}{1}}$: \textcolor{cyan}{1-Dim} Value weights \\
      \sout{$W_o^h \in \mathbb{R}^{d_h \times d}$} $\textcolor{cyan}{w_o^h} \in \mathbb{R}^{\textcolor{cyan}{1} \times d}$: \textcolor{cyan}{1-Dim} Output weights \\
      $H \in \mathbb{Z^+}$: Number of Lorsa heads \\
      $\textcolor{cyan}{K} \in \mathbb{Z^+}$: \textcolor{cyan}{Max number of activated Lorsa Heads}
  }
  \KwOut{$\mathbf{\hat{Y}} \in \mathbb{R}^{n \times d}$: Output sequence}
  
  \For{$h \leftarrow 1$ \textbf{to} $H$}{
      $Q^h = X W_q^h \in \mathbb{R}^{n \times d_h}$ \tcp*{Query projection for head $h$}
      $K^h = X W_k^h \in \mathbb{R}^{n \times d_h}$ \tcp*{Key projection}
      $\textcolor{cyan}{v^h} = X w_v^h \in \mathbb{R}^{n \times \textcolor{cyan}{1}}$ \tcp*{\sout{$d_h$-Dim} \textcolor{cyan}{1-Dim} Value projection}
      
      $A^h = \text{softmax}\left(\frac{Q^h (K^h)^T}{\sqrt{d_h}}\right) \in \mathbb{R}^{n \times n}$ \tcp*{Attention patterns}
      $\textcolor{cyan}{z^h} = A^h v^h \in \mathbb{R}^{n \times \textcolor{cyan}{1}}$ \tcp*{\sout{$d_h$-Dim} \textcolor{cyan}{1-Dimensional} Weighted sum of values}
      $\mathbf{\hat{Y}^h} = z^h w_o^h \in \mathbb{R}^{n \times d}$ \tcp*{Output of a single Lorsa head}
  }
  
  $\textcolor{cyan}{\mathcal{S} \leftarrow \text{TopKIndices}(\{z^h \mid h=1,\dots,H\}, K)}$ \tcp*{\textcolor{cyan}{Select top K heads by $z$}}
  $\mathbf{\hat{Y}} = \sum_{h \textcolor{cyan}{\in \mathcal{S}}} \mathbf{\hat{Y}^h}$ \tcp*{Combine \sout{all} \textcolor{cyan}{selected} heads}
  
  \Return{$\mathbf{\hat{Y}}$}
\end{algorithm}

We detail Lorsa's architectural designs in this section, with Algorithm~\ref{alg:lorsa} highlighting how Lorsa
architecture differs from a standard MHSA. Lorsa takes in the same inputs of MHSA and is trained
to predict MHSA outputs. The training objective is simply minimizing the mean square error (MSE):
$\mathcal{L} = \mathbb{E}_{\textbf{x}\in \mathcal{D}}||\text{Lorsa}(\textbf{x}) - \text{MHSA}(\textbf{x})||_2$.

\paragraph{One-Dimensional Output and Values.}
Each MHSA head reads from and writes to a residual stream subspace via its OV circuit~\citep{elhage2021framework}, whose dimension is decided by its head
dimension $d_h$. Under the linear representation hypothesis that unidimensional features are encoded in the residual stream, we design
Lorsa heads with 1D OV circuits. This offers the advantage of restricting read/write operations to one or few residual stream features (directions).
Although ideal implementations would use 1D QK and OV circuits, we restrict dimensionality reduction to OV circuits for practical reasons.

\paragraph{Query and Key Weights with Parameter Sharing.}

We observe significant performance drop as $D_{\text{QK}}^\text{Lorsa}$ decreases, which is severer when
$D_\text{QK}^\text{Lorsa} < D_{\text{QK}}^{\text{MHSA}}$. This may suggest QK circuits for attention units are multidimensional.
In result, we choose
$D_{\text{QK}}^{\text{Lorsa}} = D_{\text{QK}}^{\text{MHSA}}$ and implement parameter sharing for QK weights across every
$D_{\text{QK}}^{\text{Lorsa}}$ heads as the default setting. This strategy maintains a parameter count of $4D_\text{model}$ per head - equivalent to
setting $D_{\text{QK}}^{\text{Lorsa}}$ to 1 without parameter sharing, which is crucial for Lorsa scalability. 

Our parameter binding strategy renders Lorsa QK circuit strikingly similar to MHSA - a QK-sharing group of Lorsa heads is almost identical to
an original MHSA head except the sparsity constraints applied on each OV dimension. We describe Lorsa heads as individual heads with shared QK
circuits rather than a sparse dimension in MHSA architecture because they often exhibit correlated yet distinct interpretable functionalities,
as we will show in Section~\ref{sec:search_specific_lorsa_heads}. And there are cases where a QK-sharing group of Lorsa heads show no clear
semantic correlation. We also show that Lorsa QK circuits are not solely learning to copy of the original QK circuit as shown in Appendix~\ref{appendix:ablation:init}.
This distinguishes Lorsa from only applying sparse dictionary learning or Independent Component Analysis on OV circuits~\citep{ameisen2024ovica}.

\paragraph{Orders of Magnitudes More Heads and Sparsity.} 
To capture numerous underlying attention units, Lorsa employs an overcomplete architecture with $N_\text{Lorsa}\gg N_\text{MHSA}$ heads per layer,
activating only $K\ll N_\text{Lorsa}$ heads per token. This parallels
Sparse Autoencoders' approach of learning more features than the input dimension while enforcing sparsity.

For a given token position, Lorsa's output aggregates the Top-K heads with largest $z$'s, where $z$ is the
scalar activation value of a Lorsa head\footnote{Conceptually, a Lorsa head's activation on a sequence should be $z^h ||w_o^h||_2$ rather than $z^h$. For analytic simplicity
and clarity, we construct a model with identical predictions but set $w_v^h\leftarrow w_v^h||w_o^h||_2$, $b_v^h \leftarrow b_v^h||w_o^h||_2$ and
$w_o^h\leftarrow w_o^h / ||w_o^h||_2$. This operation isolates activation $z^h$ from output direction $w_o^h$.}. The active head subset dynamically
varies across token positions. This sparsity mechanism resembles TopK-SAEs~\citep{gao2024oaisae}, as both select the $K$ most salient linear components.

\paragraph{Connection to Sparse Autoencoders.}
Lorsa shows notable resemblance to attention SAEs~\citep{kissane2024attentionsae} for its 1D OV circuits. Lorsa learns an overcomplete
linear basis of the attention output space $\{w_o^h \mid h=1,...,H\}$ with sparsely activated scalar components $\{z^h_i \mid h=1,...,H\}$ at the $i$-th position,
which is analogous to SAE decoder and sparse feature activations.

However, whereas SAE features are computed via single linear encoders with ReLU, Lorsa head activation as a given position $z^h_i$ derives from attention
patterns $A^h_i$ and $v^h$ of previous tokens. Moreover, SAEs take in and predict the same activations while Lorsa, like Transcoders~\citep{ge2024hierattr,dunefsky2024transcoder}
, learns to predict downstream activations. Lorsa is similar to a Gated~\citep{rajamanoharan2024gatedsae} Transcoder taking in
activations from multiple positions, where the QK circuit resembles the \emph{gate} for its non-linearity and $w_v$ is simply a linear encoder.

\subsection{Lorsa Training}
The Low-Rank Sparse Attention modules we are studying throughout this work are trained on all layers of Pythia-160M
and Llama-3.1-8B. The training set is sampled from 800 million tokens for each model, which is adequate to train Lorsa models with till
convergence. The prompts are collected from SlimPajama~\citep{cerebras2023slimpajama} truncated to a context size of 256 for Pythia
and 1024 for Llama. We report our experimental settings in Table~\ref{tab:trained_lorsa_overview}. Appendix~\ref{appendix:lnk_scaling_law} details lorsa training
settings along with a Lorsa $L(N,K)$ scaling law compared against TopK SAEs. 

\begin{table}[h]
  \centering
  \adjustbox{width=\linewidth}{
    \begin{tabular}{@{}c|cccc|ccc|cc|cc@{}}
      \toprule
      \multirow{2}{*}{Target Model} &
        \multicolumn{4}{c|}{\# Heads} &
        \multicolumn{3}{c|}{Head Dimension} &
        \multicolumn{2}{c|}{\begin{tabular}[c]{@{}c@{}}\# Active Heads\\ per Token\end{tabular}} &
        \multicolumn{2}{c}{\begin{tabular}[c]{@{}c@{}}\# Params\\ Per Layer\end{tabular}} \\ \cmidrule(l){2-12} &
        \multicolumn{1}{c|}{MHSA} &
        \multicolumn{1}{c|}{\begin{tabular}[c]{@{}c@{}}Independent \\ Lorsa QK\end{tabular}} &
        \multicolumn{1}{c|}{\begin{tabular}[c]{@{}c@{}}Lorsa\\ QK\end{tabular}} &
        \begin{tabular}[c]{@{}c@{}}Lorsa\\ OV\end{tabular} &
        \multicolumn{1}{c|}{MHSA} &
        \multicolumn{1}{c|}{\begin{tabular}[c]{@{}c@{}}Lorsa\\ QK\end{tabular}} &
        \begin{tabular}[c]{@{}c@{}}Lorsa\\ OV\end{tabular} &
        \multicolumn{1}{c|}{MHSA} &
        Lorsa &
        \multicolumn{1}{c|}{MHSA} &
        Lorsa \\ \midrule
      Pythia-160M &
        12 &
        96 &
        6K &
        6K &
        64 &
        64 &
        1 &
        12 &
        64 &
        2.25M &
        18M \\
      Llama-3.1-8B &
        32 &
        256 &
        32K &
        32K &
        128 &
        128 &
        1 &
        32 &
        128 &
        64M &
        512M \\ \bottomrule
      \end{tabular}
}
\caption{Experimental setups for both target models. We primarily focus on Lorsa modules with 500-1,000 times more heads than the
original MHSA. For instance, we have 6K Lorsa heads for an MHSA layer in Pythia-160M, with every
$D_{\text{QK}}^{\text{Lorsa}}=D_{\text{QK}}^{\text{MHSA}}=64$ heads sharing QK weights. This gives us 96 independent QK weights.}
\label{tab:trained_lorsa_overview}
\end{table}

\section{Assessing Lorsa Interpretability}
\label{sec:interpret_lorsa_heads}

\subsection{Interpreting Individual Lorsa Heads}
\label{sec:interpret_lorsa_heads:main_property}

\paragraph{Top Activations.}
With Lorsa heads' output restricted to a single direction, their activation strength at a given position $i$ can be described with a scalar
$z^h_i$ (Section~\ref{sec:lorsa:arch}). Similar to SAE interpretation methods~\citep{bricken2023monosemanticity,templeton2024scaling}, we
iterate over 100M activations from a held-out dataset to identify the 16 highest-activating tokens for each Lorsa head.

\paragraph{$\vect{z}$ Pattern.}

According to Algorithm~\ref{alg:lorsa}, the top activations $z_i^h$ decompose linearly into token-wise contributions from preceding positions:
$z_i^h = A_i^h v^h = \sum_{j=1}^{i}A_{i,j}^h v_j^h$, where $A_{i,j}^h$ denotes attention weight from token $i$ to token $j$ and $v_j^h = w_v^h \mathbf{x}_j$.
Conceptually this tells from which previous tokens the activation $z_i^h$ is computed. Thus we call it the $z$ pattern.
This is analogous to direct feature attribution (DFA) analysis for attention SAEs~\citep{kissane2024attentionsae,he2024othello}.
An SAE feature's activation at the $i$-th token $f_i$ can be decomposed along heads and sequence position, i.e., $f_i = \sum_{j\le i} \sum_{h\in H} W^\text{enc}_f o_j^h$, where $o_j^h$ is a linear component of
MHSA output at token $j$ from head $h$. The DFA from token $j$
is then defined as $\sum_{h\in H} W^\text{enc}_f o_j^h$. In comparison, Lorsa's attribution includes only one 1D OV circuit and a single, though shared, QK circuit without multi-head aggregation.
This enables QK circuit attribution for attention units distributed across multiple MHSA heads.

\subsection{Visualization Interface}
\label{sec:interpret_lorsa_heads:interface}
\begin{figure}[h]
  \centering
  \includegraphics[width=\textwidth]{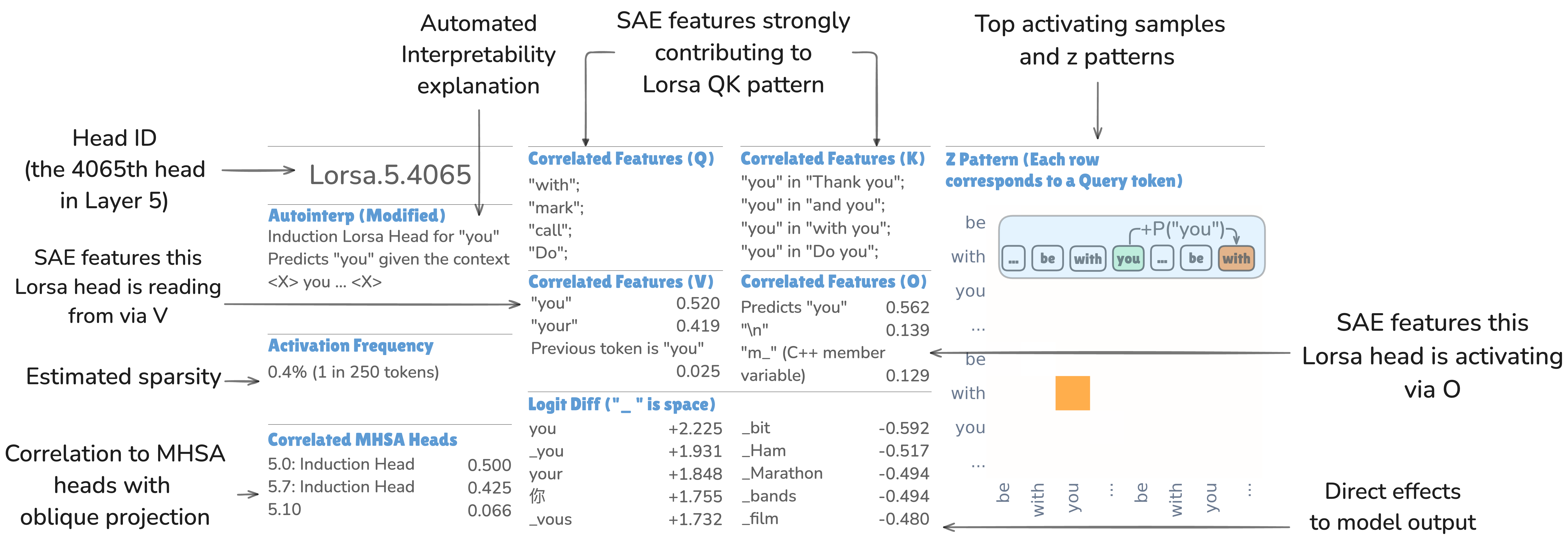}
  \caption{Visualization dashboard for a "you"-specific induction Lorsa head. We provide an example interpretation of each
  item below.}
  \label{fig:dashboards}
\end{figure}

Our visualization interface provides multifaceted information on Lorsa head interpretation.
We illustrate our dashboards with the example in Figure~\ref{fig:dashboards}, which visualizes to an induction Lorsa head
specifically firing for the token "you". The methods used to identify correlated MHSA heads and SAE features are described in
Appendix~\ref{appendix:corr_wiz_MHSA}~and~\ref{appendix:corr_wiz_SAE_features}.

\begin{itemize}[leftmargin=*]
    \item \textbf{Correlation to SAE features / Logits via OV:} It mainly reads from \emph{current token is "you"/"your"} features via its $w_v^h$;
    It strongly activates a \emph{say "you"} feature (i.e., a feature amplifying the logit of "you" via the logit lens~\citep{nostalgebraist2020logitlens});
    It amplifies the logits of a variety of "you" tokens.

    \item \textbf{Correlation to SAE features via QK:} Its QK attention pattern is mainly computed by \emph{current token is "X"} features on the query position and
    \emph{previous token is "X" \& current token is "you"} features, where "X" includes a number of tokens that often precedes "you", such as "with", "thank" or "do".

    \item \textbf{Correlation to MHSA heads:} This Lorsa head is almost equally distributed in MHSA.5.0 and MHSA.5.7. Both MHSA heads exhibit
    induction functionality as shown in Appendix~\ref{appendix:corr_wiz_MHSA}.
\end{itemize}

\subsection{Quantitative Evaluation with Automated Interpretability}

\begin{figure}[h]
  \centering
  \includegraphics[width=\textwidth]{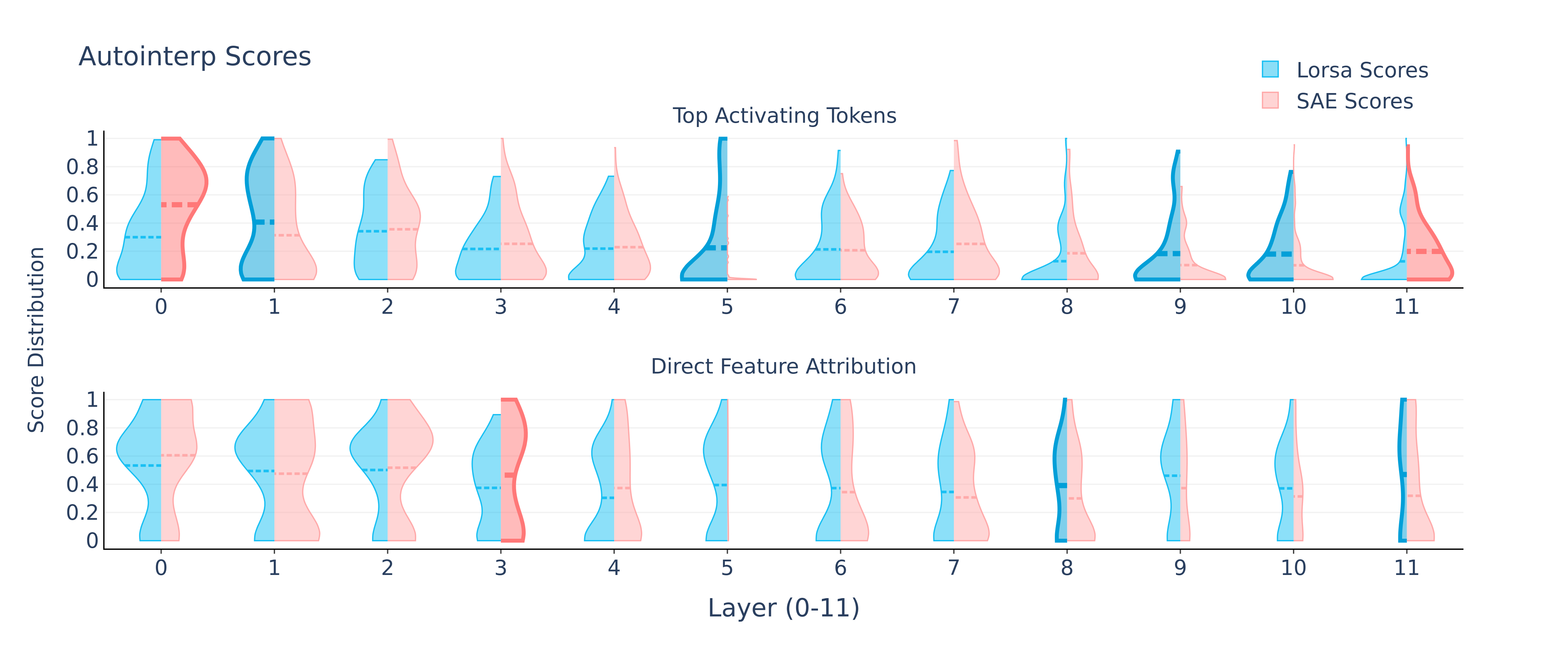}
  \caption{Automated interpretability scores of \textcolor{cyan}{Lorsa} heads and \textcolor{saepink}{SAE} features. Each distribution is estimated with 100
  heads / features. The average score of each group is represented by a horizontal dash line. We highlight distributions with larger mean value suggested by
  t-tests with $\alpha=0.05$.}
  \label{fig:autointerp_result}
\end{figure}

To quantify the interpretability of Lorsa heads in terms of its top activations and $z$ pattern, we perform automated
interpretability (autointerp)~\citep{bills2023autointerp} to estimate how comprehensible each Lorsa head is. We apply standard autointerp on max
activating samples, Lorsa $z$-patterns and direct feature attribution of attention output SAEs.
Prompt design and choice of few-shot examples are detailed in Appendix~\ref{appendix:autointerp}. All results are obtained with Pythia-160M Lorsa
and SAEs of the same size.

As shown in Figure~\ref{fig:autointerp_result}, Lorsa achieves a higher score in 6 cases, with 3 losses and 15 ties at $\alpha=0.05$ significance across 24 layer-wise comparisons,
suggesting comparable interpretability to SAE features. Both methods exhibit descending scores in deeper layers. Potential explanations include:
(1) increased polysemanticity in later layers, or (2) autointerp's limited capacity for capturing long-range dependencies.

\section{Searching for Specific Lorsa Heads}
\label{sec:search_specific_lorsa_heads}

We use path patching~\citep{wang23ioi,conmy2023ACDC} to find the Lorsa heads involved in specialized tasks. 
For a given Lorsa head, path patching ablates its output and allows the influence to propagate only through residual connections and MLPs  
(but not through other attention heads). This measures the head's counterfactual influence on the model's behavior. Using this approach, we re-discover previously
documented relatively monosemantic heads (Section~\ref{sec:search_specific_lorsa_heads:rediscover_previously_reported_heads}), identify
a family of arithmetic-specific Lorsa heads (Section~\ref{sec:search_specific_lorsa_heads:arithmetic_family_of_lorsa_heads}), and an
interesting set of \emph{broadcasting Lorsa heads} (Section~\ref{sec:search_specific_lorsa_heads:broadcasting_lorsa_heads}).

\subsection{Lorsa Re-discovers Previously Reported Heads}
\label{sec:search_specific_lorsa_heads:rediscover_previously_reported_heads}

\begin{figure}[t]
  \centering
  \includegraphics[width=\textwidth]{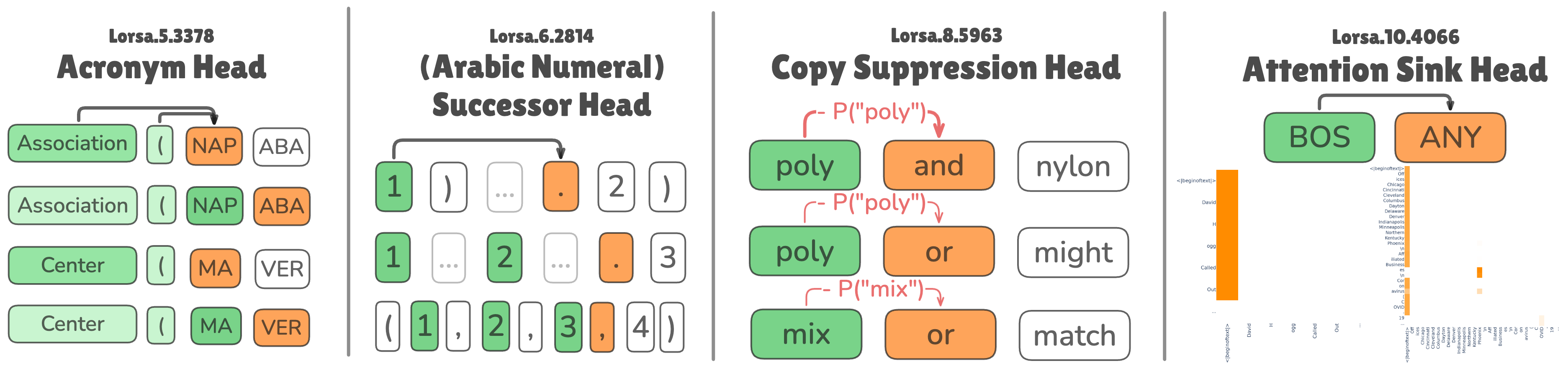}
  \caption{Examples of Lorsa heads re-discovering previously reported heads.
  \textbf{Lorsa.5.3378}: The Acronym Head attends to the parentheses and preceding text to predict the abbreviation.
  \textbf{Lorsa.6.2814}: Successor Head attends to the previous number token and predicts the next number.
  \textbf{Lorsa.8.5963}: Copy Suppression Head attends to the previous subject and suppresses its copy. 
  \textbf{Lorsa.10.4066}: Attention Sink Head simply attends to the '<|beginoftext|>' token.}
  \label{fig:headrediscovery}
\end{figure}

Previous works have documented attention heads with specific functionalities in well-characterized contexts. We demonstrate that Lorsa 
rediscovers these attention heads. Through experiments on Pythia-160M, we show that Lorsa rediscovers heads replicating these functionalities, such as
\emph{induction heads}~\citep{olsson2022inductionheads}, \emph{name mover heads}~\citep{wang23ioi}, \emph{copy suppression heads}
~\citep{mcdougall2023copysuppression}, and \emph{successor heads}~\citep{gould24successorhead}. We also observe an important attention
behavior called attention sinks~\citep{xiao2024attentionsink}.
Figure~\ref{fig:headrediscovery} showcases four such heads, with their complete information provided in 
Appendix~\ref{appendix:examples of lorsa Rediscovered Functional Heads}. A representative selection of interpretable Lorsa heads is presented in
Table~\ref{tab:lorsa_rediscovered_heads}.

\begin{minipage}{0.47\textwidth}
  \centering
  \resizebox{\textwidth}{!}{
    \begin{tabular}{|c|c|}
      \hline
      \textbf{Lorsa Head ID} & \textbf{Autointerp (Function)}  \\
      \hline
      \rowcolor{induction} Lorsa.5.3955 & Induction for "ve"\\
      \rowcolor{induction} Lorsa.5.4010 & Induction for last names \\
      \rowcolor{induction} Lorsa.7.4203 & Induction for abbreviations \\
      \rowcolor{induction} Lorsa.9.132 & Induction after "and"/"with" \\
      \hline
      \rowcolor{specific} Lorsa.4.32 & "define"/"include" in PHP\\
      \rowcolor{specific} Lorsa.4.3013 & "public static" in Java\\
      \rowcolor{specific} Lorsa.5.4035 & Say "Four"/"Five"\\
      \rowcolor{specific} Lorsa.8.142 & Apple Inc. and products (iPhone etc.) \\
      \hline
      \rowcolor{previous} Lorsa.4.5167 & Previous token is "can"/"could"\\
      \rowcolor{previous} Lorsa.11.6084 & Previous token is "make" \\
      \hline
      \rowcolor{acronym} Lorsa.4.487 & Abbreviations (parentheses/quotes) \\
      \rowcolor{acronym} Lorsa.6.1491 & Abbreviations in parentheses \\
      \rowcolor{acronym} Lorsa.6.1787 & Abbreviations in parentheses \\
      \rowcolor{acronym} Lorsa.6.5499 & Abbreviations in parentheses\\
      \hline
      \rowcolor{foreign} Lorsa.4.1420 & Russian words\\
      \rowcolor{foreign} Lorsa.9.1622 & Induction in Italian \\
      \hline
      \rowcolor{sink} Lorsa.4.4388 & Attention sinks \\
      \rowcolor{sink} Lorsa.7.862 & Attention sinks \\
      \hline
      \rowcolor{other} Lorsa.6.2592 & "the other"/"another" \\
      \rowcolor{other} Lorsa.10.1232 & Year of birth and death \\
      \hline
    \end{tabular}
  }
  \captionof{table}{A non-exhaustive collection of interpretable Lorsa heads we have found, which are grouped by color from top to bottom: 
  {\color[RGB]{30,60,180}induction heads},
  {\color[RGB]{120,0,180}specific token heads},
  {\color[RGB]{220,100,0}previous token heads},
  {\color[RGB]{0,150,0}acronym heads},
  {\color[RGB]{180,180,0}language-specific heads},
  {\color[RGB]{200,0,100}attention sink heads},
  and {\color[RGB]{0,180,180}miscellaneous heads}.}
  \label{tab:lorsa_rediscovered_heads}
\end{minipage}  
\hfill
\begin{minipage}{0.50\textwidth}
  \includegraphics[width=\textwidth]{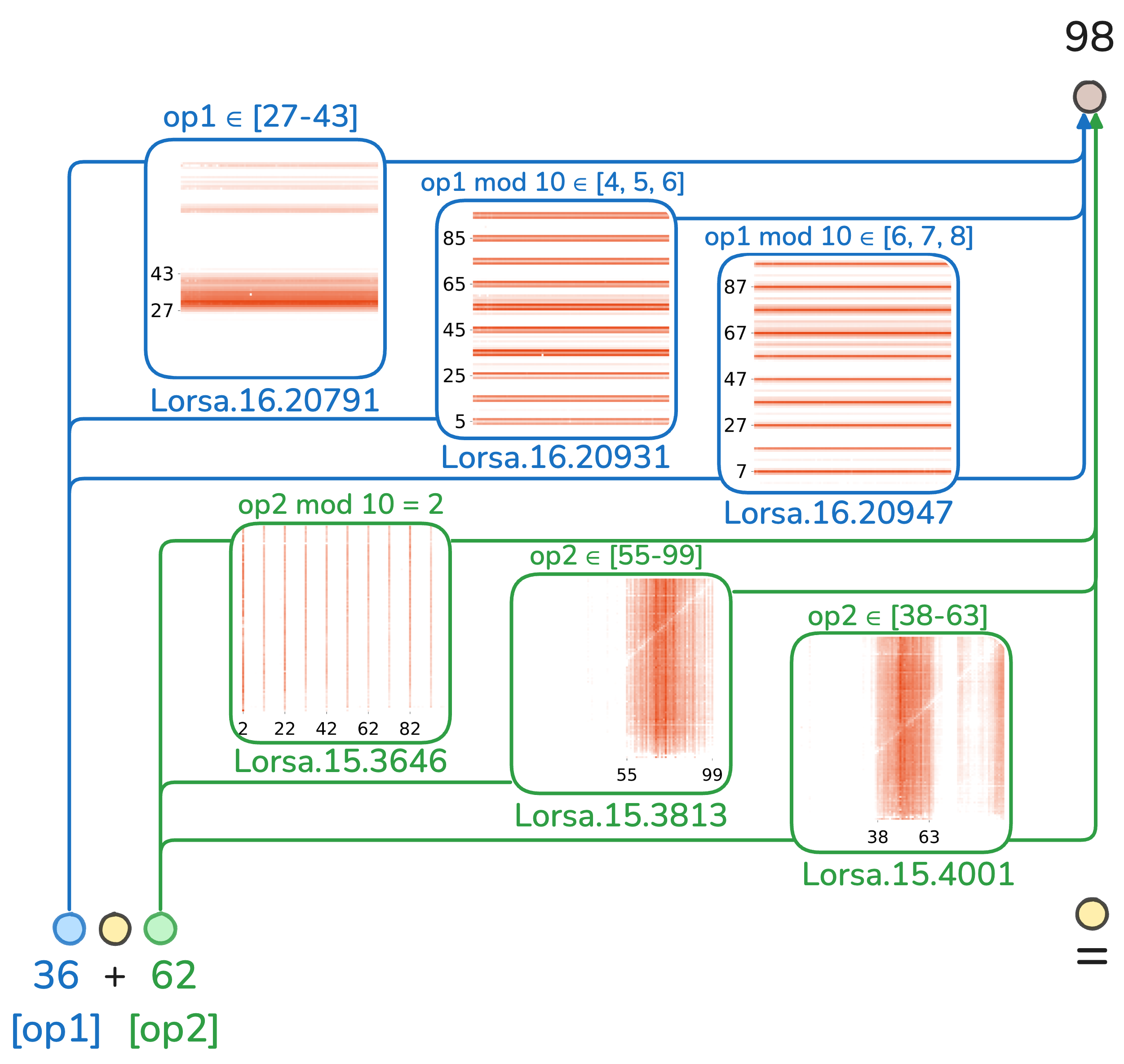}
  \captionof{figure}{For the prompt "$36+62=$", Lorsa moves two operands to the last position with 3 heads each. The first operand (36)
  is attended in terms of $z$ pattern by an "\texttt{op1} $\in \numrange{27}{43}$", an "\texttt{op1} $\%\ 10 \in [4, 5, 6]$"
  and an "\texttt{op1} $\%\ 10 \in [6, 7, 8]$" head, which uniquely determines "\texttt{op1} = 36". The same applies to \texttt{op2}.}
  \label{fig:arithmetic_case}
\end{minipage}

\subsection{A Family of Arithmetic Lorsa Heads in Llama-3.1-8B}
\label{sec:search_specific_lorsa_heads:arithmetic_family_of_lorsa_heads}

We identify a group of arithmetic-specific Lorsa heads in Llama-3.1-8B that activate during simple arithmetic operations following
the template \texttt{[op1][operator][op2][=]}.
One observation is that each head fetches certain operands with a number of unrelated heuristics,
consistent to prior findings at neuron level on arithmetic mechanisms~\citep{nikankin2024heurisrics},
despite Lorsa's architectural differences.

Figure~\ref{fig:arithmetic_case} demonstrates an example of the prompt "$36+62=$". Similar to~\citet{ameisen2025circuit}, we visualize the function of each Lorsa
head with an operand plot, displaying its activity on the 100 $\times$ 100 grid of potential inputs of the template "\texttt{op1+op2=}".

Visualization dashboards are provided in Appendix~\ref{appendix:case:arithmetic} for these six heads to support claims made in this section,
along with the more arithmetic Lorsa heads and their explanations. We also observe striking similarity between the heuristics
used by Lorsa and SAE.

\subsection{Lorsa Heads as Thematic Anchors}
\label{sec:search_specific_lorsa_heads:broadcasting_lorsa_heads}
\begin{minipage}{0.47\textwidth}
While exploring through Lorsa heads in Llama-3.1-8B, we notice a distinctive subset of Lorsa heads attending to keywords with remarkable thematic consistency
from all subsequent tokens in a sentence. Figure~\ref{fig:presidency_broadcast} illustrates a representative case which exhibit relatively
selective, long-range attention to tokens related to presidency as evidenced by $z$ pattern. Through manual inspection we also
find Lorsa heads activating on topics like alcohol addiction, dynamic system, medication instructions and terms of service.\\

An intuitive hypothetical function of these head is \emph{thematic anchors} to
maintain persistent topic representations to bias subsequent token predictions toward domain-appropriate vocabulary
and syntactic structures. We believe these heads to be closely related to SAE features "smeared" across token positions,
as mentioned in~\citet{lindsey2025biology} (\href{https://transformer-circuits.pub/2025/attribution-graphs/biology.html#structure}{link})
(\href{https://www.neuronpedia.org/llama3.1-8b/15-llamascope-res-32k/23}{example}).

\end{minipage}  
\hfill
\begin{minipage}{0.50\textwidth}
  \includegraphics[width=\textwidth]{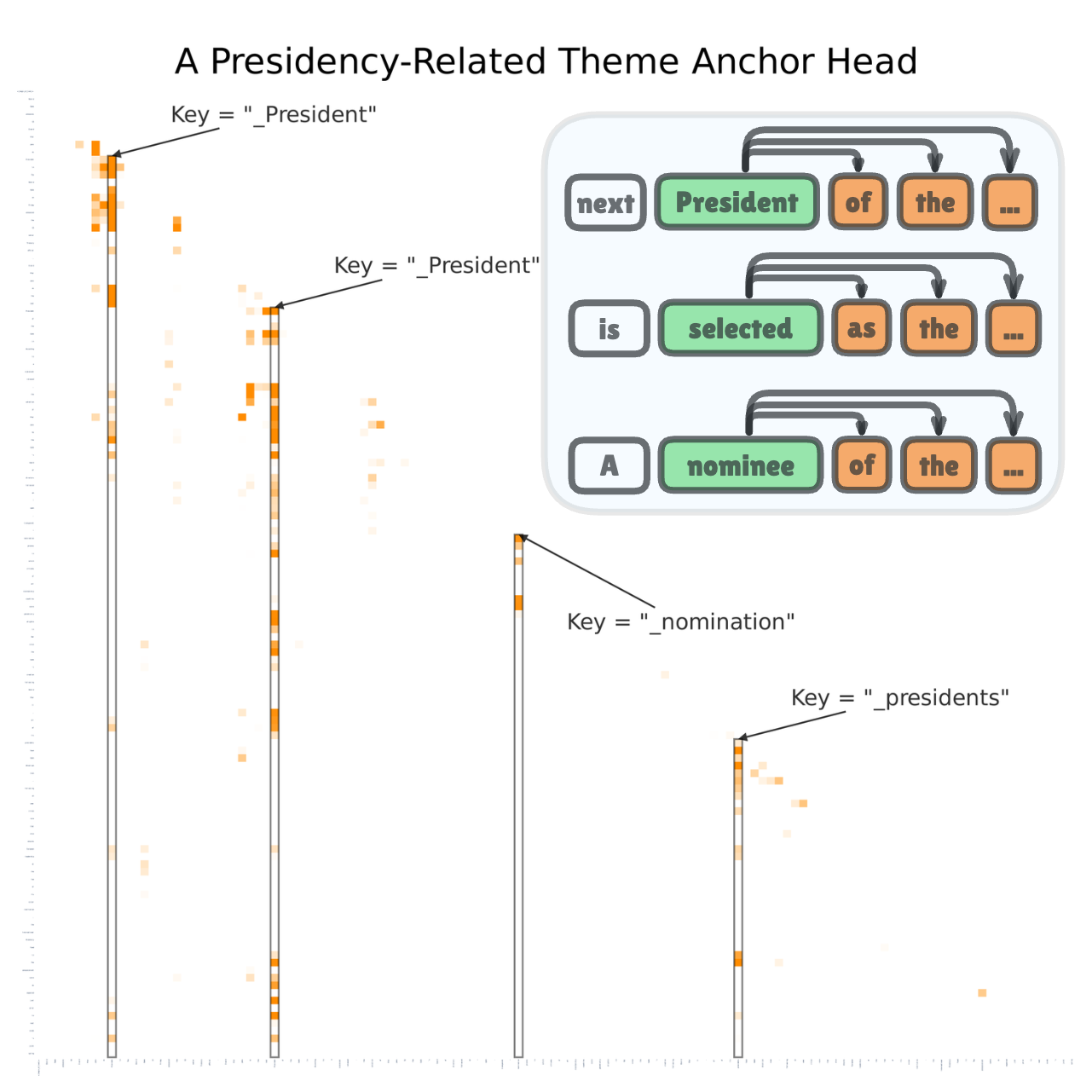}
  \captionof{figure}{$z$ pattern of a presidency-related topic broadcasting Lorsa head.}
  \label{fig:presidency_broadcast}
\end{minipage}

\section{Discussion and Limitations}
\label{sec:discussion}

We report a number of intriguing findings and limitations of Low-Rank Sparse Attention. We believe there remains significant room for improvement
for future work in each of these following aspects.

\paragraph{Reducing QK Dimension and Unbinding QK circuits.}
One significant limitation of our approach is that we do not get completely independent or low rank Lorsa heads. The shared QK circuit
of Lorsa heads raise concerns on whether they can be independently understood, despite our current positive findings with $z$ patterns, which
is an artifact of Q, K and V. If we could overcome the performance degradation of low-dimensional QK circuits, it is possible to
scale up Lorsa with more independent QK circuits and fewer residual stream features interacting via QK\footnote{It might also be the case that attention units
must be described in multidimensional QK circuits, like induction heads requiring attending to multiple "previous token is X" features.}. This is also crucial
for circuit tracing methods to clearly understand QK circuits.

\paragraph{Dark Matters.}
We find non-trivial correlation between Lorsa error and SAE errors trained on the same attention layer in terms of
(1) average loss per layer (2) loss per token on the same context and (3) error direction, as shown in Appendix~\ref{appendix:dark_matter}.
This may suggest the existence of universal dark matters~\citep{olah2024nextfivehurdles,engels2024darkmatter} for
sparse dictionary learning methods like SAE and Lorsa. Any progress along this direction to reduce or understand SAE / Lorsa
dark matters should reveal many interesting behaviors of neural networks.

\paragraph{Inactive Attention SAE Features and Lorsa Heads.}
Despite efforts on hyperparameter search, we find that attention SAE and Lorsa both contains a majority of inactive feature / heads.
This phenomenon renders most computation wasted and raises a question about the difference between structure
of attention output space and MLP output space or residual streams, where SAEs of the same size only have
few dead features if configured properly.

\paragraph{Cross Layer Attention Superposition.}
If certain inter-token feature interaction is performed in more than one layer, our current method which decomposes only one MHSA layer
does not suffice to find such relation. This parallels the problem of cross-layer superposition~\citep{templeton2024scaling} for residual
stream features. A cross-layer variant of Lorsa~\citep{lindsey2024crosscoder} might be tractable.

\paragraph{Global Weights and Systematic Q/K/V Composition.}
To better understand the global attention behavior of Transformers, one important research direction is to identify systematic Q/K/V composition
like induction heads and previous token heads. Since Lorsa reveals finer-grained versions of MHSA heads, we can expect to find more of such
cross-layer collaboration behavior. However, we failed in our early attempts to find Lorsa heads with Q/K composition.

\section{Related Work}
\label{sec:related_work}

\subsection{Explaining Individual Attention Heads} 
\label{sec:related_work:interp_individual_heads}

With the help of activation patching~\citep{meng22rome,zhang24patching} or path patching~\citep{wang23ioi,conmy2023ACDC}, the literature has
discovered a number of heads that exhibit certain functionality in pre-defined contexts. This line of research starts from a composition of
\emph{previous token heads} and \emph{induction heads}~\citep{olsson2022inductionheads} which is closely related to in context learning.
More work on this line includes \emph{name mover heads}~\citep{wang23ioi}, \emph{number comparison heads}~\citep{hanna23greaterthan},
\emph{copy suppression heads}~\citep{mcdougall2023copysuppression}, \emph{successor heads}~\citep{gould24successorhead} and
\emph{long context retrieval heads}~\citep{wu24retrievalhead}.

\subsection{Superposition Hypothesis and Sparse Autoencoders}
\label{sec:related_work:sae}

The superposition hypothesis~\citep{arora2018superposition,olah2020zoom,elhage2022tms} assumes that neurons are related to multiple
non-orthognal underlying features. Sparse Autoencoders~\citep{cunningham2023sae,bricken2023monosemanticity} are proposed to extract
an overcomplete set of the sparse and linear comprehensible features. Importantly, the success of the technique also sheds light on
universality of superposition across model size~\citep{templeton2024scaling,lieberum2024gemmascope,he2024llamascope}, model
architectures~\citep{wang2024univ} and modality~\citep{abdulaal2024xray}.

\subsection{Sparse Autoencoder Variants}
\label{sec:related_work:sae_variants}

We see SAEs to have developed multiple forms along with the rapid evolution of SAEs in the past year. Some of them improve
initialization~\citep{conerly2024aprilhowwetrain}, loss function~\citep{conerly2024ghostgrad,bussmann2024matryoshkasaes} or sparsity
constraints~\citep{gao2024oaisae} to solve specific issues such as shrinkage~\citep{wright2024featuresupp} and massive inactive
features~\citep{bricken2023monosemanticity}.

Another direction of improvement is the SAE architecture. For instance, Gated SAEs~\citep{rajamanoharan2024gatedsae} are proved effective
in mitigating shrinkage. Transcoders~\citep{ge2024hierattr,dunefsky2024transcoder} aims to simplify sparse circuit analysis by replacing
MLPs, whose non-linear nature makes causal attribution intractable.

\section{Conclusion}
In this work, we introduced Low-Rank Sparse Attention (Lorsa) to disentangle atomic attention units from attention superposition in Transformer
models. Our experiments validated that Lorsa can recover known attention mechanisms (e.g., induction heads, name movers) and uncover new interpretable
behaviors (e.g., arithmetic-specific heads). While Lorsa improves attention interpretability, key challenges remain, particularly in unbinding QK circuits to
achieve fully independent heads and reducing superposition effects. Future work should explore low-dimensional QK structures, cross-layer superposition,
and systematic Q/K/V composition to further decompose attention mechanisms. Addressing these limitations could enable a complete sparse, interpretable
reconstruction of Transformer computations, advancing our understanding of in-context learning and feature interactions.

\newpage
\bibliography{neurips_2024}

\begin{thebibliography}{51}
\providecommand{\natexlab}[1]{#1}
\providecommand{\url}[1]{\texttt{#1}}
\expandafter\ifx\csname urlstyle\endcsname\relax
  \providecommand{\doi}[1]{doi: #1}\else
  \providecommand{\doi}{doi: \begingroup \urlstyle{rm}\Url}\fi

\bibitem[Abdulaal et~al.(2024)Abdulaal, Fry, Brown, Ijishakin, Gao, Hyland, Alexander, and Castro]{abdulaal2024xray}
Ahmed Abdulaal, Hugo Fry, Nina~Monta{\~{n}}a Brown, Ayodeji Ijishakin, Jack Gao, Stephanie~L. Hyland, Daniel~C. Alexander, and Daniel~C. Castro.
\newblock An x-ray is worth 15 features: Sparse autoencoders for interpretable radiology report generation.
\newblock \emph{CoRR}, abs/2410.03334, 2024.
\newblock \doi{10.48550/ARXIV.2410.03334}.
\newblock URL \url{https://doi.org/10.48550/arXiv.2410.03334}.

\bibitem[Ainslie et~al.(2023)Ainslie, Lee{-}Thorp, de~Jong, Zemlyanskiy, Lebr{\'{o}}n, and Sanghai]{ainslie2023gqa}
Joshua Ainslie, James Lee{-}Thorp, Michiel de~Jong, Yury Zemlyanskiy, Federico Lebr{\'{o}}n, and Sumit Sanghai.
\newblock {GQA:} training generalized multi-query transformer models from multi-head checkpoints.
\newblock In Houda Bouamor, Juan Pino, and Kalika Bali, editors, \emph{Proceedings of the 2023 Conference on Empirical Methods in Natural Language Processing, {EMNLP} 2023, Singapore, December 6-10, 2023}, pages 4895--4901. Association for Computational Linguistics, 2023.
\newblock \doi{10.18653/V1/2023.EMNLP-MAIN.298}.
\newblock URL \url{https://doi.org/10.18653/v1/2023.emnlp-main.298}.

\bibitem[Ameisen et~al.(2024)Ameisen, Batson, and Lindsey]{ameisen2024ovica}
Emmanuel Ameisen, Joshua Batson, and Jack Lindsey.
\newblock Investigating successor heads.
\newblock \emph{Transformer Circuits Thread}, 2024.
\newblock URL \url{https://transformer-circuits.pub/2024/september-update/index.html}.

\bibitem[Ameisen et~al.(2025)Ameisen, Lindsey, Pearce, Gurnee, Turner, Chen, Citro, Abrahams, Carter, Hosmer, Marcus, Sklar, Templeton, Bricken, McDougall, Cunningham, Henighan, Jermyn, Jones, Persic, Qi, Ben~Thompson, Zimmerman, Rivoire, Conerly, Olah, and Batson]{ameisen2025circuit}
Emmanuel Ameisen, Jack Lindsey, Adam Pearce, Wes Gurnee, Nicholas~L. Turner, Brian Chen, Craig Citro, David Abrahams, Shan Carter, Basil Hosmer, Jonathan Marcus, Michael Sklar, Adly Templeton, Trenton Bricken, Callum McDougall, Hoagy Cunningham, Thomas Henighan, Adam Jermyn, Andy Jones, Andrew Persic, Zhenyi Qi, T.~Ben~Thompson, Sam Zimmerman, Kelley Rivoire, Thomas Conerly, Chris Olah, and Joshua Batson.
\newblock Circuit tracing: Revealing computational graphs in language models.
\newblock \emph{Transformer Circuits Thread}, 2025.
\newblock URL \url{https://transformer-circuits.pub/2025/attribution-graphs/methods.html}.

\bibitem[Arora et~al.(2018)Arora, Li, Liang, Ma, and Risteski]{arora2018superposition}
Sanjeev Arora, Yuanzhi Li, Yingyu Liang, Tengyu Ma, and Andrej Risteski.
\newblock Linear algebraic structure of word senses, with applications to polysemy.
\newblock \emph{Trans. Assoc. Comput. Linguistics}, 6:\penalty0 483--495, 2018.
\newblock \doi{10.1162/TACL\_A\_00034}.
\newblock URL \url{https://doi.org/10.1162/tacl\_a\_00034}.

\bibitem[Batson et~al.(2024)Batson, Chen, and Jones]{batson2024attribution}
Joshua Batson, Brian Chen, and Andy Jones.
\newblock Circuits updates - march 2024.
\newblock \emph{Transformer Circuits Thread}, 2024.
\newblock URL \url{https://transformer-circuits.pub/2024/march-update/index.html}.

\bibitem[Biderman et~al.(2023)Biderman, Schoelkopf, Anthony, Bradley, O'Brien, Hallahan, Khan, Purohit, Prashanth, Raff, Skowron, Sutawika, and van~der Wal]{Biderman2023Pythia}
Stella Biderman, Hailey Schoelkopf, Quentin~Gregory Anthony, Herbie Bradley, Kyle O'Brien, Eric Hallahan, Mohammad~Aflah Khan, Shivanshu Purohit, USVSN~Sai Prashanth, Edward Raff, Aviya Skowron, Lintang Sutawika, and Oskar van~der Wal.
\newblock Pythia: {A} suite for analyzing large language models across training and scaling.
\newblock In Andreas Krause, Emma Brunskill, Kyunghyun Cho, Barbara Engelhardt, Sivan Sabato, and Jonathan Scarlett, editors, \emph{International Conference on Machine Learning, {ICML} 2023, 23-29 July 2023, Honolulu, Hawaii, {USA}}, volume 202 of \emph{Proceedings of Machine Learning Research}, pages 2397--2430. {PMLR}, 2023.
\newblock URL \url{https://proceedings.mlr.press/v202/biderman23a.html}.

\bibitem[Bills et~al.(2023)Bills, Cammarata, Mossing, Tillman, Gao, Goh, Sutskever, Leike, Wu, and Saunders]{bills2023autointerp}
Steven Bills, Nick Cammarata, Dan Mossing, Henk Tillman, Leo Gao, Gabriel Goh, Ilya Sutskever, Jan Leike, Jeff Wu, and William Saunders.
\newblock Language models can explain neurons in language models.
\newblock \url{https://openaipublic.blob.core.windows.net/neuron-explainer/paper/index.html}, 2023.

\bibitem[Bricken et~al.(2023)Bricken, Templeton, Batson, Chen, Jermyn, Conerly, Turner, Anil, Denison, Askell, Lasenby, Wu, Kravec, Schiefer, Maxwell, Joseph, Hatfield-Dodds, Tamkin, Nguyen, McLean, Burke, Hume, Carter, Henighan, and Olah]{bricken2023monosemanticity}
Trenton Bricken, Adly Templeton, Joshua Batson, Brian Chen, Adam Jermyn, Tom Conerly, Nick Turner, Cem Anil, Carson Denison, Amanda Askell, Robert Lasenby, Yifan Wu, Shauna Kravec, Nicholas Schiefer, Tim Maxwell, Nicholas Joseph, Zac Hatfield-Dodds, Alex Tamkin, Karina Nguyen, Brayden McLean, Josiah~E Burke, Tristan Hume, Shan Carter, Tom Henighan, and Christopher Olah.
\newblock Towards monosemanticity: Decomposing language models with dictionary learning.
\newblock \emph{Transformer Circuits Thread}, 2023.
\newblock URL \url{https://transformer-circuits.pub/2023/monosemantic-features/index.html}.

\bibitem[Bussmann et~al.(2024)Bussmann, Leask, and Nanda]{bussmann2024matryoshkasaes}
Bart Bussmann, Patrick Leask, and Neel Nanda.
\newblock Learning multi-level features with matryoshka saes.
\newblock LessWrong, 2024.
\newblock URL \url{https://www.lesswrong.com/posts/rKM9b6B2LqwSB5ToN/learning-multi-level-features-with-matryoshka-saes}.

\bibitem[Conerly(2024)]{conerly2024ghostgrad}
Tom Conerly.
\newblock Circuits updates - february 2024.
\newblock \emph{Transformer Circuits Thread}, 2024.
\newblock URL \url{https://transformer-circuits.pub/2024/feb-update/index.html#dict-learning-resampling}.

\bibitem[Conerly et~al.(2024)Conerly, Templeton, Bricken, Marcus, and Henighan]{conerly2024aprilhowwetrain}
Tom Conerly, Adly Templeton, Trenton Bricken, Jonathan Marcus, and Tom Henighan.
\newblock Circuits updates - april 2024.
\newblock \emph{Transformer Circuits Thread}, 2024.
\newblock URL \url{https://transformer-circuits.pub/2024/april-update/index.html#training-saes}.

\bibitem[Conmy et~al.(2023)Conmy, Mavor{-}Parker, Lynch, Heimersheim, and Garriga{-}Alonso]{conmy2023ACDC}
Arthur Conmy, Augustine~N. Mavor{-}Parker, Aengus Lynch, Stefan Heimersheim, and Adri{\`{a}} Garriga{-}Alonso.
\newblock Towards automated circuit discovery for mechanistic interpretability.
\newblock In Alice Oh, Tristan Naumann, Amir Globerson, Kate Saenko, Moritz Hardt, and Sergey Levine, editors, \emph{Advances in Neural Information Processing Systems 36: Annual Conference on Neural Information Processing Systems 2023, NeurIPS 2023, New Orleans, LA, USA, December 10 - 16, 2023}, 2023.
\newblock URL \url{http://papers.nips.cc/paper\_files/paper/2023/hash/34e1dbe95d34d7ebaf99b9bcaeb5b2be-Abstract-Conference.html}.

\bibitem[Cunningham et~al.(2023)Cunningham, Ewart, Riggs, Huben, and Sharkey]{cunningham2023sae}
Hoagy Cunningham, Aidan Ewart, Logan Riggs, Robert Huben, and Lee Sharkey.
\newblock Sparse autoencoders find highly interpretable features in language models.
\newblock \emph{CoRR}, abs/2309.08600, 2023.
\newblock \doi{10.48550/ARXIV.2309.08600}.
\newblock URL \url{https://doi.org/10.48550/arXiv.2309.08600}.

\bibitem[Dubey et~al.(2024)Dubey, Jauhri, Pandey, Kadian, Al{-}Dahle, Letman, Mathur, Schelten, Yang, Fan, Goyal, Hartshorn, Yang, Mitra, Sravankumar, Korenev, Hinsvark, Rao, Zhang, Rodriguez, Gregerson, Spataru, Rozi{\`{e}}re, Biron, Tang, Chern, Caucheteux, Nayak, Bi, Marra, McConnell, Keller, Touret, Wu, Wong, Ferrer, Nikolaidis, Allonsius, Song, Pintz, Livshits, Esiobu, Choudhary, Mahajan, Garcia{-}Olano, Perino, Hupkes, Lakomkin, AlBadawy, Lobanova, Dinan, Smith, Radenovic, Zhang, Synnaeve, Lee, Anderson, Nail, Mialon, Pang, Cucurell, Nguyen, Korevaar, Xu, Touvron, Zarov, Ibarra, Kloumann, Misra, Evtimov, Copet, Lee, Geffert, Vranes, Park, Mahadeokar, Shah, van~der Linde, Billock, Hong, Lee, Fu, Chi, Huang, Liu, Wang, Yu, Bitton, Spisak, Park, Rocca, Johnstun, Saxe, Jia, Alwala, Upasani, Plawiak, Li, Heafield, Stone, and et~al.]{dubey2024llama3}
Abhimanyu Dubey, Abhinav Jauhri, Abhinav Pandey, Abhishek Kadian, Ahmad Al{-}Dahle, Aiesha Letman, Akhil Mathur, Alan Schelten, Amy Yang, Angela Fan, Anirudh Goyal, Anthony Hartshorn, Aobo Yang, Archi Mitra, Archie Sravankumar, Artem Korenev, Arthur Hinsvark, Arun Rao, Aston Zhang, Aur{\'{e}}lien Rodriguez, Austen Gregerson, Ava Spataru, Baptiste Rozi{\`{e}}re, Bethany Biron, Binh Tang, Bobbie Chern, Charlotte Caucheteux, Chaya Nayak, Chloe Bi, Chris Marra, Chris McConnell, Christian Keller, Christophe Touret, Chunyang Wu, Corinne Wong, Cristian~Canton Ferrer, Cyrus Nikolaidis, Damien Allonsius, Daniel Song, Danielle Pintz, Danny Livshits, David Esiobu, Dhruv Choudhary, Dhruv Mahajan, Diego Garcia{-}Olano, Diego Perino, Dieuwke Hupkes, Egor Lakomkin, Ehab AlBadawy, Elina Lobanova, Emily Dinan, Eric~Michael Smith, Filip Radenovic, Frank Zhang, Gabriel Synnaeve, Gabrielle Lee, Georgia~Lewis Anderson, Graeme Nail, Gr{\'{e}}goire Mialon, Guan Pang, Guillem Cucurell, Hailey Nguyen, Hannah Korevaar, Hu~Xu, Hugo Touvron, Iliyan Zarov, Imanol~Arrieta Ibarra, Isabel~M. Kloumann, Ishan Misra, Ivan Evtimov, Jade Copet, Jaewon Lee, Jan Geffert, Jana Vranes, Jason Park, Jay Mahadeokar, Jeet Shah, Jelmer van~der Linde, Jennifer Billock, Jenny Hong, Jenya Lee, Jeremy Fu, Jianfeng Chi, Jianyu Huang, Jiawen Liu, Jie Wang, Jiecao Yu, Joanna Bitton, Joe Spisak, Jongsoo Park, Joseph Rocca, Joshua Johnstun, Joshua Saxe, Junteng Jia, Kalyan~Vasuden Alwala, Kartikeya Upasani, Kate Plawiak, Ke~Li, Kenneth Heafield, Kevin Stone, and et~al.
\newblock The llama 3 herd of models.
\newblock \emph{CoRR}, abs/2407.21783, 2024.
\newblock \doi{10.48550/ARXIV.2407.21783}.
\newblock URL \url{https://doi.org/10.48550/arXiv.2407.21783}.

\bibitem[Dunefsky et~al.(2024)Dunefsky, Chlenski, and Nanda]{dunefsky2024transcoder}
Jacob Dunefsky, Philippe Chlenski, and Neel Nanda.
\newblock Transcoders find interpretable {LLM} feature circuits.
\newblock \emph{CoRR}, abs/2406.11944, 2024.
\newblock \doi{10.48550/ARXIV.2406.11944}.
\newblock URL \url{https://doi.org/10.48550/arXiv.2406.11944}.

\bibitem[Elhage et~al.(2021)Elhage, Nanda, Olsson, Henighan, Joseph, Mann, Askell, Bai, Chen, Conerly, DasSarma, Drain, Ganguli, Hatfield-Dodds, Hernandez, Jones, Kernion, Lovitt, Ndousse, Amodei, Brown, Clark, Kaplan, McCandlish, and Olah]{elhage2021framework}
Nelson Elhage, Neel Nanda, Catherine Olsson, Tom Henighan, Nicholas Joseph, Ben Mann, Amanda Askell, Yuntao Bai, Anna Chen, Tom Conerly, Nova DasSarma, Dawn Drain, Deep Ganguli, Zac Hatfield-Dodds, Danny Hernandez, Andy Jones, Jackson Kernion, Liane Lovitt, Kamal Ndousse, Dario Amodei, Tom Brown, Jack Clark, Jared Kaplan, Sam McCandlish, and Chris Olah.
\newblock A mathematical framework for transformer circuits.
\newblock \emph{Transformer Circuits Thread}, 2021.
\newblock https://transformer-circuits.pub/2021/framework/index.html.

\bibitem[Elhage et~al.(2022)Elhage, Hume, Olsson, Schiefer, Henighan, Kravec, Hatfield-Dodds, Lasenby, Drain, Chen, Grosse, McCandlish, Kaplan, Amodei, Wattenberg, and Olah]{elhage2022tms}
Nelson Elhage, Tristan Hume, Catherine Olsson, Nicholas Schiefer, Tom Henighan, Shauna Kravec, Zac Hatfield-Dodds, Robert Lasenby, Dawn Drain, Carol Chen, Roger Grosse, Sam McCandlish, Jared Kaplan, Dario Amodei, Martin Wattenberg, and Christopher Olah.
\newblock Toy models of superposition.
\newblock \emph{Transformer Circuits Thread}, 2022.
\newblock URL \url{https://transformer-circuits.pub/2022/toy\_model/index.html}.

\bibitem[Elhage et~al.(2023)Elhage, Lasenby, and Olah]{elhage2023privbasis}
Nelson Elhage, Robert Lasenby, and Christopher Olah.
\newblock Privileged bases in the transformer residual stream.
\newblock \emph{Transformer Circuits Thread}, 2023.
\newblock URL \url{https://transformer-circuits.pub/2023/privileged-basis/index.html}.

\bibitem[Engels et~al.(2024)Engels, Riggs, and Tegmark]{engels2024darkmatter}
Joshua Engels, Logan Riggs, and Max Tegmark.
\newblock Decomposing the dark matter of sparse autoencoders.
\newblock \emph{CoRR}, abs/2410.14670, 2024.
\newblock \doi{10.48550/ARXIV.2410.14670}.
\newblock URL \url{https://doi.org/10.48550/arXiv.2410.14670}.

\bibitem[Gao et~al.(2024)Gao, la~Tour, Tillman, Goh, Troll, Radford, Sutskever, Leike, and Wu]{gao2024oaisae}
Leo Gao, Tom~Dupr{\'{e}} la~Tour, Henk Tillman, Gabriel Goh, Rajan Troll, Alec Radford, Ilya Sutskever, Jan Leike, and Jeffrey Wu.
\newblock Scaling and evaluating sparse autoencoders.
\newblock \emph{CoRR}, abs/2406.04093, 2024.
\newblock \doi{10.48550/ARXIV.2406.04093}.
\newblock URL \url{https://doi.org/10.48550/arXiv.2406.04093}.

\bibitem[Ge et~al.(2024)Ge, Zhu, Shu, Wang, He, and Qiu]{ge2024hierattr}
Xuyang Ge, Fukang Zhu, Wentao Shu, Junxuan Wang, Zhengfu He, and Xipeng Qiu.
\newblock Automatically identifying local and global circuits with linear computation graphs.
\newblock \emph{CoRR}, abs/2405.13868, 2024.
\newblock \doi{10.48550/ARXIV.2405.13868}.
\newblock URL \url{https://doi.org/10.48550/arXiv.2405.13868}.

\bibitem[Gould et~al.(2024)Gould, Ong, Ogden, and Conmy]{gould24successorhead}
Rhys Gould, Euan Ong, George Ogden, and Arthur Conmy.
\newblock Successor heads: Recurring, interpretable attention heads in the wild.
\newblock In \emph{The Twelfth International Conference on Learning Representations, {ICLR} 2024, Vienna, Austria, May 7-11, 2024}. OpenReview.net, 2024.
\newblock URL \url{https://openreview.net/forum?id=kvcbV8KQsi}.

\bibitem[Gurnee et~al.(2023)Gurnee, Nanda, Pauly, Harvey, Troitskii, and Bertsimas]{gurnee23neurons_haystack}
Wes Gurnee, Neel Nanda, Matthew Pauly, Katherine Harvey, Dmitrii Troitskii, and Dimitris Bertsimas.
\newblock Finding neurons in a haystack: Case studies with sparse probing.
\newblock \emph{Trans. Mach. Learn. Res.}, 2023, 2023.
\newblock URL \url{https://openreview.net/forum?id=JYs1R9IMJr}.

\bibitem[Hanna et~al.(2023)Hanna, Liu, and Variengien]{hanna23greaterthan}
Michael Hanna, Ollie Liu, and Alexandre Variengien.
\newblock How does {GPT-2} compute greater-than?: Interpreting mathematical abilities in a pre-trained language model.
\newblock In Alice Oh, Tristan Naumann, Amir Globerson, Kate Saenko, Moritz Hardt, and Sergey Levine, editors, \emph{Advances in Neural Information Processing Systems 36: Annual Conference on Neural Information Processing Systems 2023, NeurIPS 2023, New Orleans, LA, USA, December 10 - 16, 2023}, 2023.
\newblock URL \url{http://papers.nips.cc/paper\_files/paper/2023/hash/efbba7719cc5172d175240f24be11280-Abstract-Conference.html}.

\bibitem[He et~al.(2024{\natexlab{a}})He, Ge, Tang, Sun, Cheng, and Qiu]{he2024othello}
Zhengfu He, Xuyang Ge, Qiong Tang, Tianxiang Sun, Qinyuan Cheng, and Xipeng Qiu.
\newblock Dictionary learning improves patch-free circuit discovery in mechanistic interpretability: {A} case study on othello-gpt.
\newblock \emph{CoRR}, abs/2402.12201, 2024{\natexlab{a}}.
\newblock \doi{10.48550/ARXIV.2402.12201}.
\newblock URL \url{https://doi.org/10.48550/arXiv.2402.12201}.

\bibitem[He et~al.(2024{\natexlab{b}})He, Shu, Ge, Chen, Wang, Zhou, Liu, Guo, Huang, Wu, Jiang, and Qiu]{he2024llamascope}
Zhengfu He, Wentao Shu, Xuyang Ge, Lingjie Chen, Junxuan Wang, Yunhua Zhou, Frances Liu, Qipeng Guo, Xuanjing Huang, Zuxuan Wu, Yu{-}Gang Jiang, and Xipeng Qiu.
\newblock Llama scope: Extracting millions of features from llama-3.1-8b with sparse autoencoders.
\newblock \emph{CoRR}, abs/2410.20526, 2024{\natexlab{b}}.
\newblock \doi{10.48550/ARXIV.2410.20526}.
\newblock URL \url{https://doi.org/10.48550/arXiv.2410.20526}.

\bibitem[Jermyn et~al.(2024)Jermyn, Olah, and Conerly]{jermyn2024attentionsuperposition}
Adam Jermyn, Chris Olah, and Tom Conerly.
\newblock Circuits updates - january 2024.
\newblock \emph{Transformer Circuits Thread}, 2024.
\newblock URL \url{https://transformer-circuits.pub/2024/jan-update/index.html#attn-superposition}.

\bibitem[Kissane et~al.(2024)Kissane, Krzyzanowski, Bloom, Conmy, and Nanda]{kissane2024attentionsae}
Connor Kissane, Robert Krzyzanowski, Joseph~Isaac Bloom, Arthur Conmy, and Neel Nanda.
\newblock Interpreting attention layer outputs with sparse autoencoders.
\newblock \emph{CoRR}, abs/2406.17759, 2024.
\newblock \doi{10.48550/ARXIV.2406.17759}.
\newblock URL \url{https://doi.org/10.48550/arXiv.2406.17759}.

\bibitem[Krzyzanowski et~al.(2024)Krzyzanowski, Kissane, Conmy, and Nanda]{krzyzanowski2024inspect_every}
Robert Krzyzanowski, Connor Kissane, Arthur Conmy, and Neel Nanda.
\newblock We inspected every head in gpt-2 small using saes so you don’t have to.
\newblock Alignment Forum, 2024.
\newblock URL \url{https://www.alignmentforum.org/posts/xmegeW5mqiBsvoaim/we-inspected-every-head-in-gpt-2-small-using-saes-so-you-don}.

\bibitem[Lieberum et~al.(2024)Lieberum, Rajamanoharan, Conmy, Smith, Sonnerat, Varma, Kram{\'{a}}r, Dragan, Shah, and Nanda]{lieberum2024gemmascope}
Tom Lieberum, Senthooran Rajamanoharan, Arthur Conmy, Lewis Smith, Nicolas Sonnerat, Vikrant Varma, J{\'{a}}nos Kram{\'{a}}r, Anca~D. Dragan, Rohin Shah, and Neel Nanda.
\newblock Gemma scope: Open sparse autoencoders everywhere all at once on gemma 2.
\newblock \emph{CoRR}, abs/2408.05147, 2024.
\newblock \doi{10.48550/ARXIV.2408.05147}.
\newblock URL \url{https://doi.org/10.48550/arXiv.2408.05147}.

\bibitem[Lindsey et~al.(2024)Lindsey, Templeton, Marcus, Conerly, Batson, and Olah]{lindsey2024crosscoder}
Jack Lindsey, Adly Templeton, Jonathan Marcus, Thomas Conerly, Joshua Batson, and Christopher Olah.
\newblock Sparse crosscoders for cross-layer features and model diffing.
\newblock \emph{Transformer Circuits Thread}, 2024.
\newblock URL \url{https://transformer-circuits.pub/2024/crosscoders/index.html}.

\bibitem[Lindsey et~al.(2025)Lindsey, Gurnee, Ameisen, Chen, Pearce, Turner, Citro, Abrahams, Carter, Hosmer, Marcus, Sklar, Templeton, Bricken, McDougall, Cunningham, Henighan, Jermyn, Jones, Persic, Qi, Thompson, Zimmerman, Rivoire, Conerly, Olah, and Batson]{lindsey2025biology}
Jack Lindsey, Wes Gurnee, Emmanuel Ameisen, Brian Chen, Adam Pearce, Nicholas~L. Turner, Craig Citro, David Abrahams, Shan Carter, Basil Hosmer, Jonathan Marcus, Michael Sklar, Adly Templeton, Trenton Bricken, Callum McDougall, Hoagy Cunningham, Thomas Henighan, Adam Jermyn, Andy Jones, Andrew Persic, Zhenyi Qi, T.~Ben Thompson, Sam Zimmerman, Kelley Rivoire, Thomas Conerly, Chris Olah, and Joshua Batson.
\newblock On the biology of a large language model.
\newblock \emph{Transformer Circuits Thread}, 2025.
\newblock URL \url{https://transformer-circuits.pub/2025/attribution-graphs/biology.html}.

\bibitem[McDougall et~al.(2023)McDougall, Conmy, Rushing, McGrath, and Nanda]{mcdougall2023copysuppression}
Callum McDougall, Arthur Conmy, Cody Rushing, Thomas McGrath, and Neel Nanda.
\newblock Copy suppression: Comprehensively understanding an attention head.
\newblock \emph{CoRR}, abs/2310.04625, 2023.
\newblock \doi{10.48550/ARXIV.2310.04625}.
\newblock URL \url{https://doi.org/10.48550/arXiv.2310.04625}.

\bibitem[Meng et~al.(2022)Meng, Bau, Andonian, and Belinkov]{meng22rome}
Kevin Meng, David Bau, Alex Andonian, and Yonatan Belinkov.
\newblock Locating and editing factual associations in {GPT}.
\newblock In Sanmi Koyejo, S.~Mohamed, A.~Agarwal, Danielle Belgrave, K.~Cho, and A.~Oh, editors, \emph{Advances in Neural Information Processing Systems 35: Annual Conference on Neural Information Processing Systems 2022, NeurIPS 2022, New Orleans, LA, USA, November 28 - December 9, 2022}, 2022.
\newblock URL \url{http://papers.nips.cc/paper\_files/paper/2022/hash/6f1d43d5a82a37e89b0665b33bf3a182-Abstract-Conference.html}.

\bibitem[Nikankin et~al.(2024)Nikankin, Reusch, Mueller, and Belinkov]{nikankin2024heurisrics}
Yaniv Nikankin, Anja Reusch, Aaron Mueller, and Yonatan Belinkov.
\newblock Arithmetic without algorithms: Language models solve math with a bag of heuristics.
\newblock \emph{CoRR}, abs/2410.21272, 2024.
\newblock \doi{10.48550/ARXIV.2410.21272}.
\newblock URL \url{https://doi.org/10.48550/arXiv.2410.21272}.

\bibitem[nostalgebraist(2020)]{nostalgebraist2020logitlens}
nostalgebraist.
\newblock interpreting gpt: the logit lens.
\newblock lesswrong, 2020.
\newblock URL \url{https://www.lesswrong.com/posts/AcKRB8wDpdaN6v6ru/interpreting-gpt-the-logit-lens}.

\bibitem[Olah and Jermyn(2024)]{olah2024nextfivehurdles}
Chris Olah and Adam Jermyn.
\newblock Circuits updates - july 2024.
\newblock \emph{Transformer Circuits Thread}, 2024.
\newblock URL \url{https://transformer-circuits.pub/2024/july-update/index.html#hurdles}.

\bibitem[Olah et~al.(2020)Olah, Cammarata, Schubert, Goh, Petrov, and Carter]{olah2020zoom}
Chris Olah, Nick Cammarata, Ludwig Schubert, Gabriel Goh, Michael Petrov, and Shan Carter.
\newblock Zoom in: An introduction to circuits.
\newblock \emph{Distill}, 2020.
\newblock \doi{10.23915/distill.00024.001}.
\newblock https://distill.pub/2020/circuits/zoom-in.

\bibitem[Olsson et~al.(2022)Olsson, Elhage, Nanda, Joseph, DasSarma, Henighan, Mann, Askell, Bai, Chen, Conerly, Drain, Ganguli, Hatfield-Dodds, Hernandez, Johnston, Jones, Kernion, Lovitt, Ndousse, Amodei, Brown, Clark, Kaplan, McCandlish, and Olah]{olsson2022inductionheads}
Catherine Olsson, Nelson Elhage, Neel Nanda, Nicholas Joseph, Nova DasSarma, Tom Henighan, Ben Mann, Amanda Askell, Yuntao Bai, Anna Chen, Tom Conerly, Dawn Drain, Deep Ganguli, Zac Hatfield-Dodds, Danny Hernandez, Scott Johnston, Andy Jones, Jackson Kernion, Liane Lovitt, Kamal Ndousse, Dario Amodei, Tom Brown, Jack Clark, Jared Kaplan, Sam McCandlish, and Chris Olah.
\newblock In-context learning and induction heads.
\newblock \emph{Transformer Circuits Thread}, 2022.
\newblock URL \url{https://transformer-circuits.pub/2022/in-context-learning-and-induction-heads/index.html}.

\bibitem[Rajamanoharan et~al.(2024)Rajamanoharan, Conmy, Smith, Lieberum, Varma, Kram{\'{a}}r, Shah, and Nanda]{rajamanoharan2024gatedsae}
Senthooran Rajamanoharan, Arthur Conmy, Lewis Smith, Tom Lieberum, Vikrant Varma, J{\'{a}}nos Kram{\'{a}}r, Rohin Shah, and Neel Nanda.
\newblock Improving dictionary learning with gated sparse autoencoders.
\newblock \emph{CoRR}, abs/2404.16014, 2024.
\newblock \doi{10.48550/ARXIV.2404.16014}.
\newblock URL \url{https://doi.org/10.48550/arXiv.2404.16014}.

\bibitem[Soboleva et~al.(2023)Soboleva, Al-Khateeb, Myers, Steeves, Hestness, and Dey]{cerebras2023slimpajama}
Daria Soboleva, Faisal Al-Khateeb, Robert Myers, Jacob~R Steeves, Joel Hestness, and Nolan Dey.
\newblock {SlimPajama: A 627B token cleaned and deduplicated version of RedPajama}.
\newblock \url{https://cerebras.ai/blog/slimpajama-a-627b-token-cleaned-and-deduplicated-version-of-redpajama}, 2023.
\newblock URL \url{https://huggingface.co/datasets/cerebras/SlimPajama-627B}.

\bibitem[Su et~al.(2021)Su, Lu, Pan, Wen, and Liu]{su2021roformer}
Jianlin Su, Yu~Lu, Shengfeng Pan, Bo~Wen, and Yunfeng Liu.
\newblock Roformer: Enhanced transformer with rotary position embedding.
\newblock \emph{CoRR}, abs/2104.09864, 2021.
\newblock URL \url{https://arxiv.org/abs/2104.09864}.

\bibitem[Templeton et~al.(2024{\natexlab{a}})Templeton, Batson, Jermyn, and Olah]{templeton2024predictfuture}
Adly Templeton, Joshua Batson, Adam Jermyn, and Chris Olah.
\newblock Circuits updates - january 2024.
\newblock \emph{Transformer Circuits Thread}, 2024{\natexlab{a}}.
\newblock URL \url{https://transformer-circuits.pub/2024/jan-update/index.html#predict-future}.

\bibitem[Templeton et~al.(2024{\natexlab{b}})Templeton, Conerly, Marcus, Lindsey, Bricken, Chen, Pearce, Citro, Ameisen, Jones, Cunningham, Turner, McDougall, MacDiarmid, Freeman, Sumers, Rees, Batson, Jermyn, Carter, Olah, and Henighan]{templeton2024scaling}
Adly Templeton, Tom Conerly, Jonathan Marcus, Jack Lindsey, Trenton Bricken, Brian Chen, Adam Pearce, Craig Citro, Emmanuel Ameisen, Andy Jones, Hoagy Cunningham, Nicholas~L Turner, Callum McDougall, Monte MacDiarmid, C.~Daniel Freeman, Theodore~R. Sumers, Edward Rees, Joshua Batson, Adam Jermyn, Shan Carter, Chris Olah, and Tom Henighan.
\newblock Scaling monosemanticity: Extracting interpretable features from claude 3 sonnet.
\newblock \emph{Transformer Circuits Thread}, 2024{\natexlab{b}}.
\newblock URL \url{https://transformer-circuits.pub/2024/scaling-monosemanticity/index.html}.

\bibitem[Wang et~al.(2024)Wang, Ge, Shu, Tang, Zhou, He, and Qiu]{wang2024univ}
Junxuan Wang, Xuyang Ge, Wentao Shu, Qiong Tang, Yunhua Zhou, Zhengfu He, and Xipeng Qiu.
\newblock Towards universality: Studying mechanistic similarity across language model architectures.
\newblock \emph{CoRR}, abs/2410.06672, 2024.
\newblock \doi{10.48550/ARXIV.2410.06672}.
\newblock URL \url{https://doi.org/10.48550/arXiv.2410.06672}.

\bibitem[Wang et~al.(2023)Wang, Variengien, Conmy, Shlegeris, and Steinhardt]{wang23ioi}
Kevin~Ro Wang, Alexandre Variengien, Arthur Conmy, Buck Shlegeris, and Jacob Steinhardt.
\newblock Interpretability in the wild: a circuit for indirect object identification in {GPT-2} small.
\newblock In \emph{The Eleventh International Conference on Learning Representations, {ICLR} 2023, Kigali, Rwanda, May 1-5, 2023}. OpenReview.net, 2023.
\newblock URL \url{https://openreview.net/forum?id=NpsVSN6o4ul}.

\bibitem[Wright and Sharkey(2024)]{wright2024featuresupp}
Benjamin Wright and Lee Sharkey.
\newblock Addressing feature suppression in saes.
\newblock LessWrong, 2024.
\newblock URL \url{https://www.lesswrong.com/posts/3JuSjTZyMzaSeTxKk/addressing-feature-suppression-in-saes}.

\bibitem[Wu et~al.(2024)Wu, Wang, Xiao, Peng, and Fu]{wu24retrievalhead}
Wenhao Wu, Yizhong Wang, Guangxuan Xiao, Hao Peng, and Yao Fu.
\newblock Retrieval head mechanistically explains long-context factuality.
\newblock \emph{CoRR}, abs/2404.15574, 2024.
\newblock \doi{10.48550/ARXIV.2404.15574}.
\newblock URL \url{https://doi.org/10.48550/arXiv.2404.15574}.

\bibitem[Xiao et~al.(2024)Xiao, Tian, Chen, Han, and Lewis]{xiao2024attentionsink}
Guangxuan Xiao, Yuandong Tian, Beidi Chen, Song Han, and Mike Lewis.
\newblock Efficient streaming language models with attention sinks.
\newblock In \emph{The Twelfth International Conference on Learning Representations, {ICLR} 2024, Vienna, Austria, May 7-11, 2024}. OpenReview.net, 2024.
\newblock URL \url{https://openreview.net/forum?id=NG7sS51zVF}.

\bibitem[Zhang and Nanda(2024)]{zhang24patching}
Fred Zhang and Neel Nanda.
\newblock Towards best practices of activation patching in language models: Metrics and methods.
\newblock In \emph{The Twelfth International Conference on Learning Representations, {ICLR} 2024, Vienna, Austria, May 7-11, 2024}. OpenReview.net, 2024.
\newblock URL \url{https://openreview.net/forum?id=Hf17y6u9BC}.

\end{thebibliography}
\bibliographystyle{plainnat}

\newpage
\appendix

\appendixpage
\startcontents[sections]
\printcontents[sections]{l}{1}{\setcounter{tocdepth}{2}}

\newpage

\section{Applying Lorsa to MHSA Variants}
\label{appendix:MHSA_variants}

Modern transformer-based models commonly employ variants of multi-head self-attention (MHSA), such as those incorporating rotary position embeddings(RoPE)~\citep{su2021roformer} and grouped-query attention(GQA)~\citep{ainslie2023gqa}. Our proposed Lorsa method demonstrates compatibility with these MHSA variants through straightforward adaptations.

\begin{itemize}[leftmargin=*]
  \item For RoPE-enhanced MHSA, we apply the same rotary transformations to Lorsa's computed queries and keys before computing attention scores, maintaining the positional information encoding.

  \item In GQA implementations, Lorsa operates without modification—specifically, we intentionally avoid introducing grouped queries within the Lorsa framework. 
\end{itemize}

Empirical results on both Pythia-160M and LLaMA-3.1-8B demonstrate that this design choice does not adversely affect performance.

\section{Ablation Study on Crucial Architectural Designs}
\label{appendix:ablation}

We conduct ablation studies on two crucial architectural designs: (1) the query and key dimension and (2) the binding ratio. Our experiments validate the necessity of maintaining both the QK dimension and the binding mechanism in our proposed architecture. Additional ablation tests on other implementation details further validate our decisions.
  
Furthermore, we derive two \textbf{hard constraints} for parameter selection (violating these constraints leads to significant performance degradation):

\begin{itemize}[leftmargin=*]
  \item The QK dimension must not be smaller than the head dimension in MHSA
  \item The number of QK pairs must not be fewer than the number of attention heads in MHSA
\end{itemize}

\subsection{Ablation Study on QK Dimension}
\label{appendix:ablation:qk_dimension}

\begin{figure}[htbp]
  \centering
  \begin{subfigure}[b]{0.48\textwidth}
    \centering
    \includegraphics[width=\linewidth]{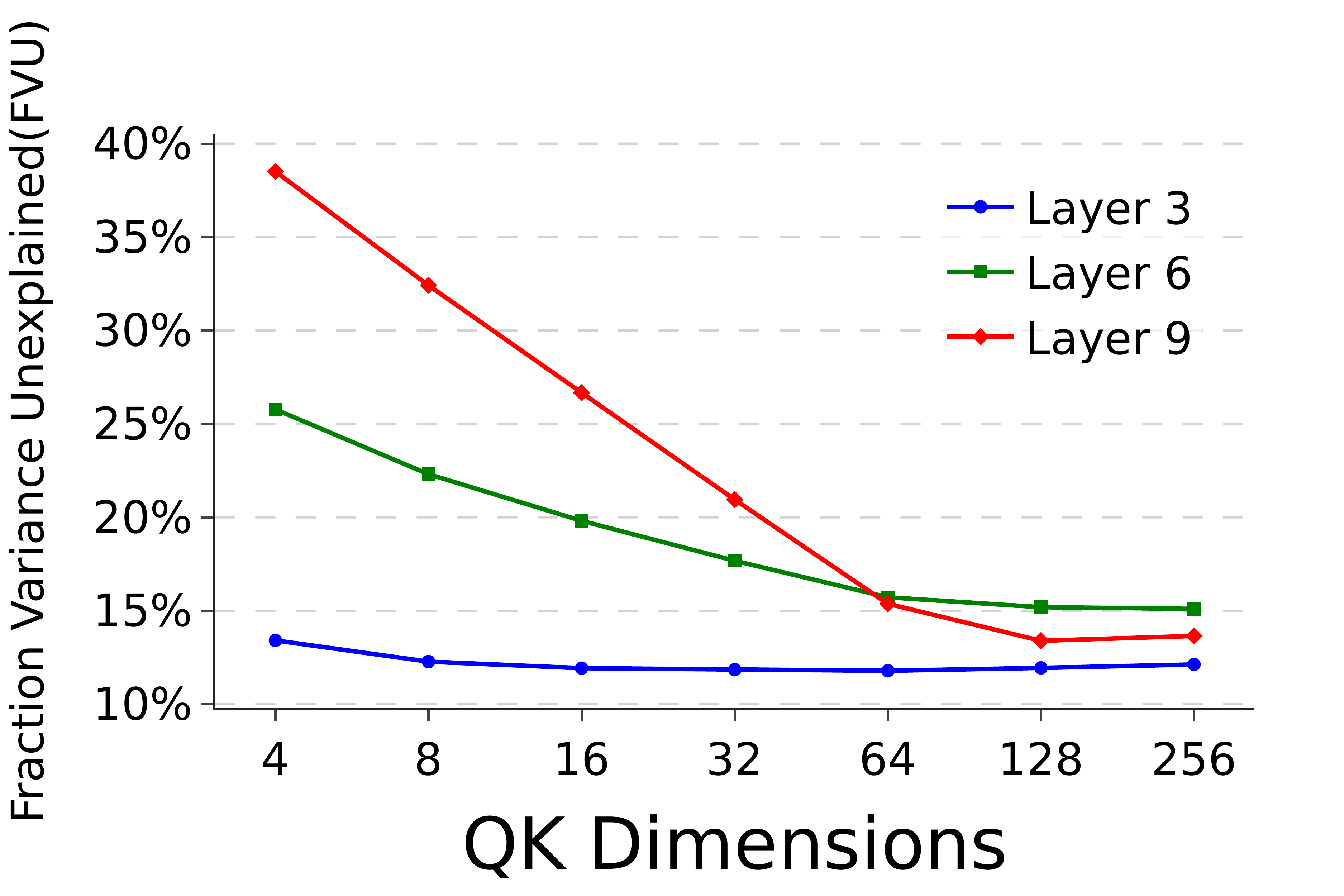}
    \caption{Context Length: 256}
    \label{fig:subfig1}
  \end{subfigure}
  \hfill
  \begin{subfigure}[b]{0.48\textwidth}
    \centering
    \includegraphics[width=\linewidth]{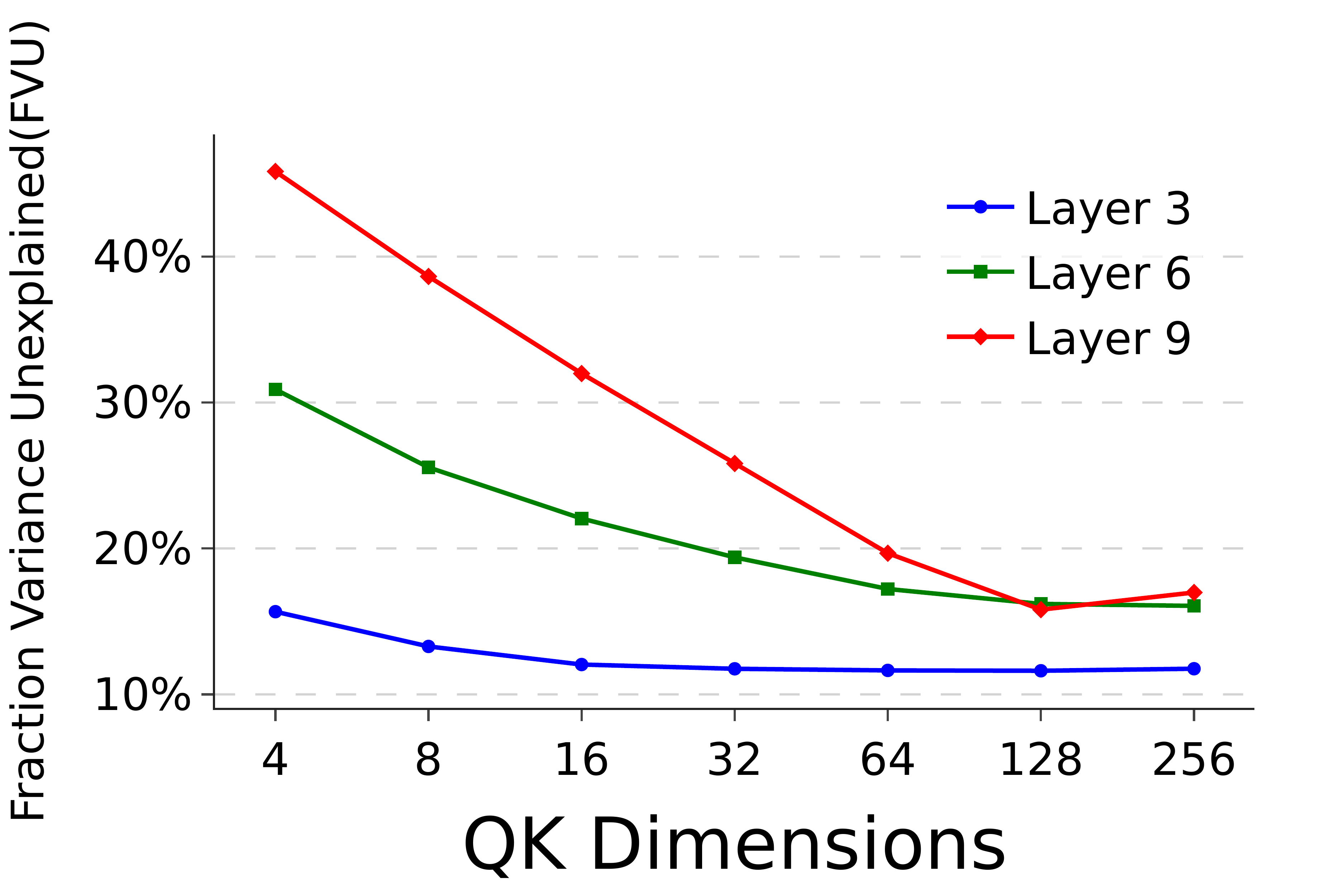}
    \caption{Context Length: 1024}
    \label{fig:subfig2}
  \end{subfigure}
  \caption{{Ablation study on the QK dimension using Pythia-160M under different context lengths. We fix the parameter budget across all settings
  and observe that reducing the QK dimension below the original MHSA head dimension (\( d_\text{head} = 64 \)) results in significant performance
  degradation, highlighting the importance of maintaining a high QK dimension.}}
  \label{fig:ablate_qk_dimension}
\end{figure}

We conduct ablation studies on the QK dimension using Pythia-160m, evaluating performance under different context lengths (256 and 1024 tokens).
To ensure fair comparison, we fix the parameter budget at 4 $D_{model}$ per attention head and maintaining a total parameter count equivalent to
$4\times$ the original MHSA configuration throughout all experiments.  As shown in Figure~\ref{fig:ablate_qk_dimension}, reducing the QK dimension below the original MHSA's head dimension ($d_\text{head}=64$) leads to severe performance degradation. This empirical evidence supports our design choice to maintain a high QK dimension.

\subsection{Ablation Study on Binding Ratio}
\label{appendix:ablation:bind_ratio}

\begin{figure}[htbp]
  \centering
  \begin{subfigure}[b]{0.32\textwidth}
    \centering
    \includegraphics[width=\linewidth]{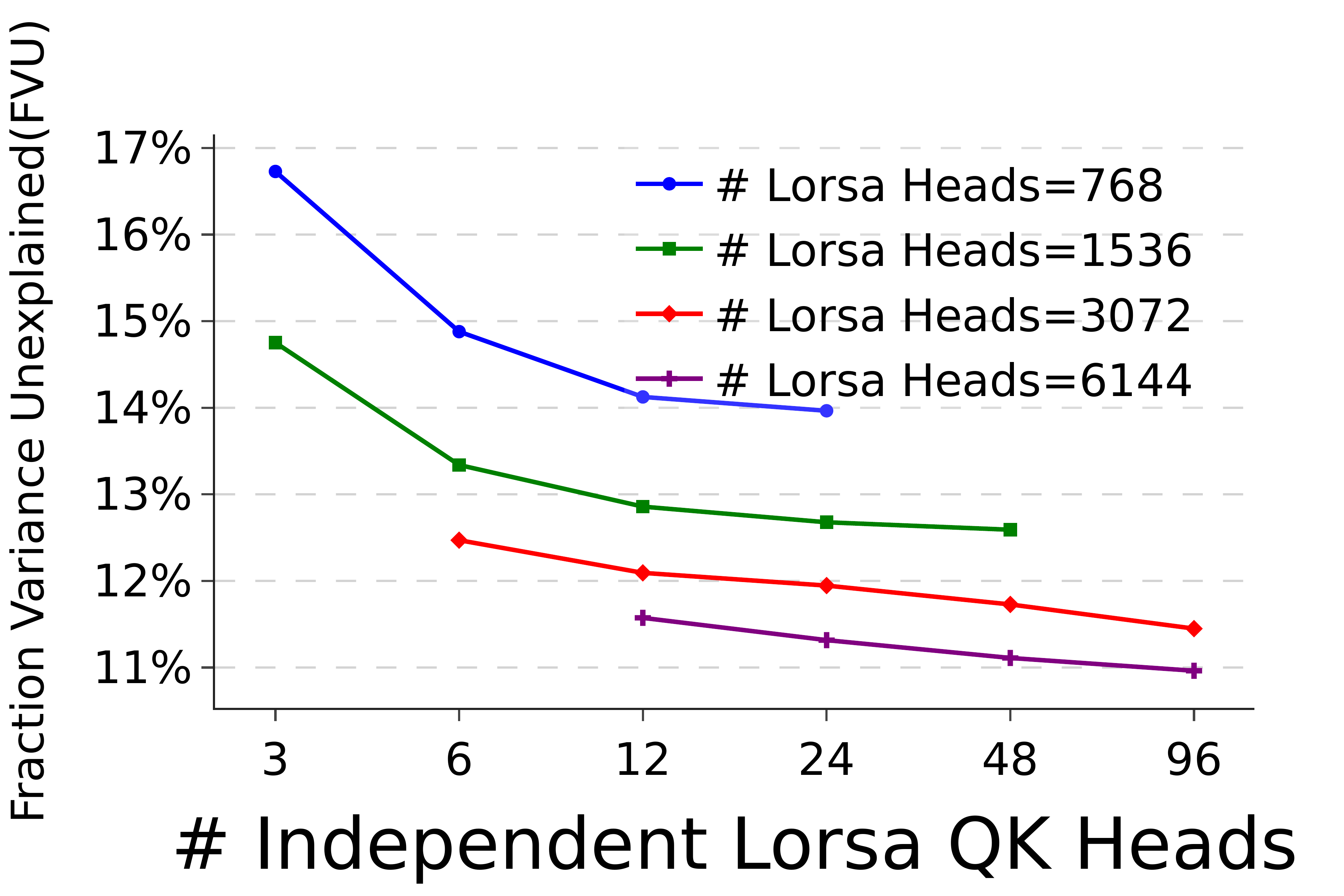}
    \caption{Context Length: 256}
    \label{fig:subfig1}
  \end{subfigure}
  \hfill
  \begin{subfigure}[b]{0.32\textwidth}
    \centering
    \includegraphics[width=\linewidth]{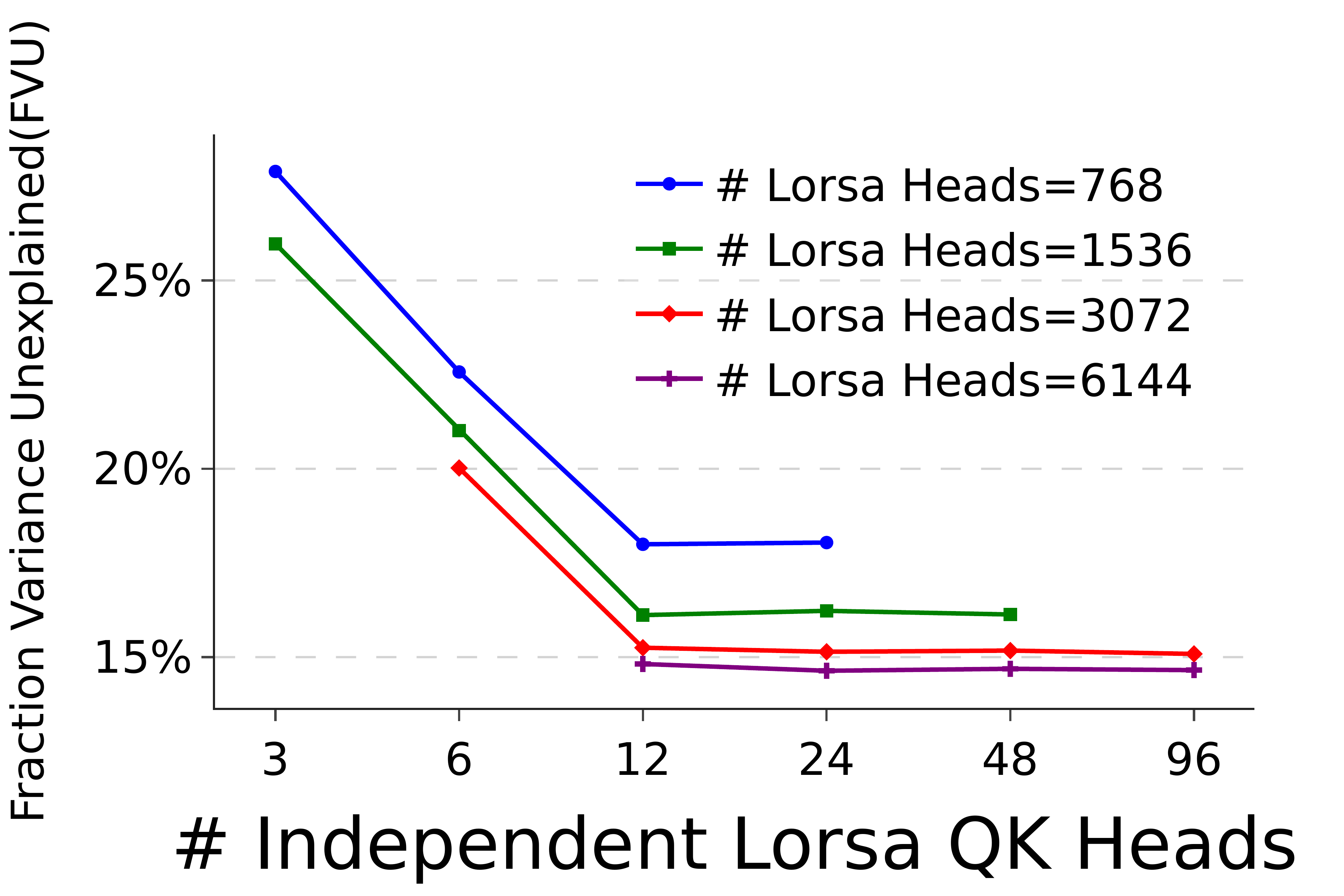}
    \caption{Context Length: 256}
    \label{fig:subfig1}
  \end{subfigure}
  \hfill
  \begin{subfigure}[b]{0.32\textwidth}
    \centering
    \includegraphics[width=\linewidth]{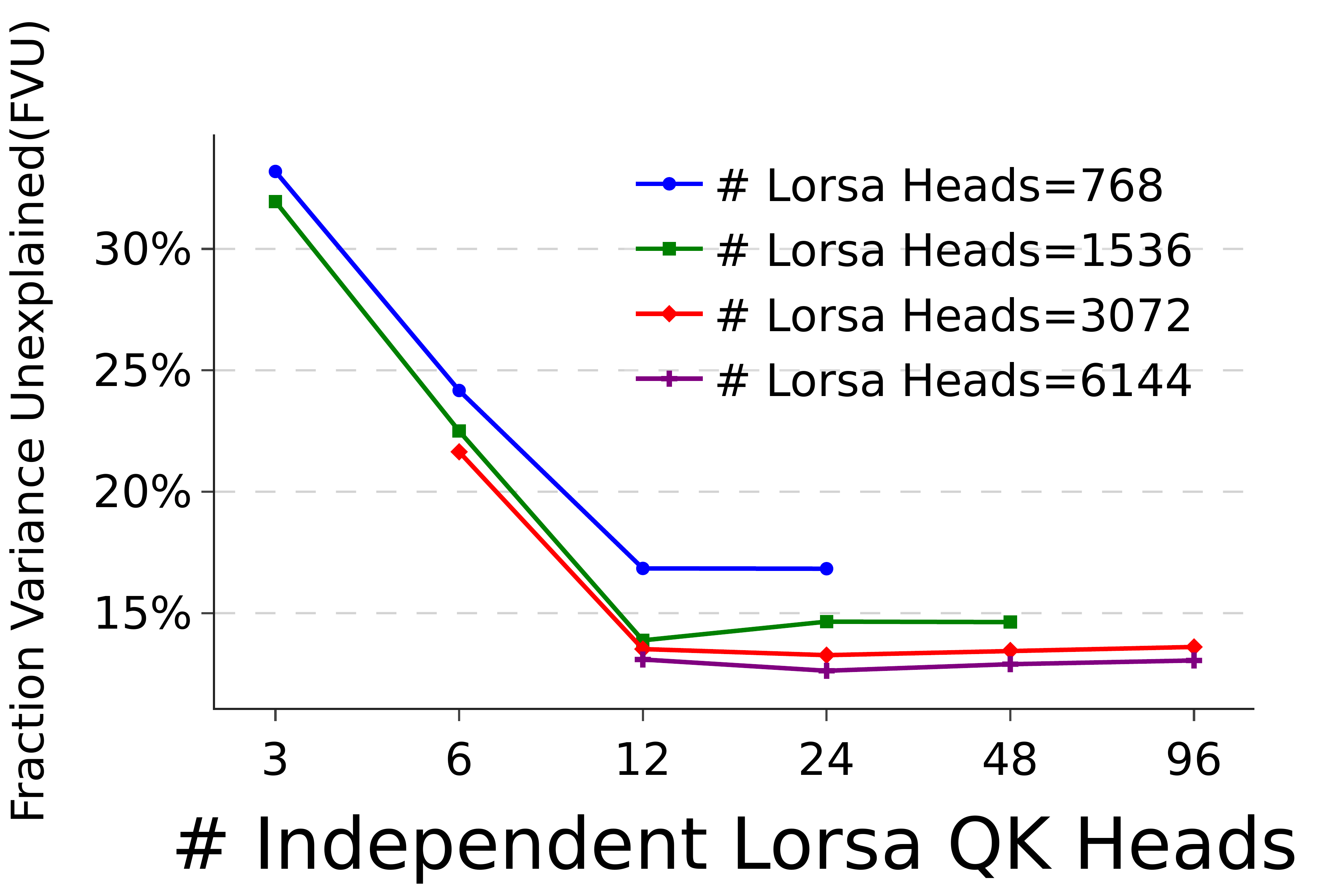}
    \caption{Context Length: 1024}
    \label{fig:subfig2}
  \end{subfigure}
  \caption{Ablation study on the binding ratio. We vary the number of independent Lorsa QK heads and evaluate model performance under different
  settings. Appropriate binding maintains performance while reducing QK circuit cost, whereas overly aggressive binding (below the number of
  original MHSA heads) leads to substantial degradation.}
  \label{fig:bind_ratio}
\end{figure}

We conduct a systematic study on the impact of the number of independent Lorsa QK heads (i.e., the number of Lorsa heads divided by the
binding ratio) across a range of configurations, as illustrated in Figure~\ref{fig:bind_ratio}. Our experimental results highlight two key observations:
\begin{itemize}[leftmargin=*]
\item Appropriate binding effectively preserves model performance while substantially reducing both the parameter count and the computational
cost of the QK circuit (scaling proportionally with the binding ratio).
\item Model performance deteriorates significantly when the number of independent QK heads falls below the original MHSA head count, establishing
this threshold as a critical lower bound for binding ratio selection.
\end{itemize}

\subsection{Ablation Study on QK Initialization}
\label{appendix:ablation:init}

Given that our QK matrices maintain high dimensionality and adopt a binding strategy, a natural question arises: can we directly reuse the original
MHSA QK parameters in Lorsa? To investigate this, we evaluate three settings: (1) randomly initializing the QK parameters of Lorsa, (2) initializing
the QK parameters of Lorsa with the original MHSA QK parameters and allowing them to be updated during training, and (3) fixing the QK parameters to
the original MHSA QK parameters throughout training. The results, summarized in Table~\ref{tab:qk_init}, show that directly fixing the QK parameters
to those of MHSA leads to worse performance compared to the other two setups. This suggests that during optimization, Lorsa learns QK parameters that
capture information not present in the original MHSA parameters.

\begin{table}[h]
\centering
\begin{tabular}{lc}
\toprule
Initialization Strategy & Fraction Variance Unexplained (FVU) \\
\midrule
Random Initialization & 11.3\% \\
Initialization with Original QK (Trainable) & 11.2\% \\
Initialization with Original QK (Fixed) & 12.4\% \\
\bottomrule
\end{tabular}
\caption{Comparison of different QK initialization strategies for Lorsa.}
\label{tab:qk_init}
\end{table}

\subsection{Implementation Details}

To align with the superposition hypothesis and the architectural design of the SAE, we apply a ReLU to ensure that the activations \( z \) are
non-negative. However, we observe that this modification has negligible impact on training dynamics, as the top-\( k \) activations are almost
always positive for reasonable choices of \( k \).

\section{$\vect{L(N, K)}$ Scaling Laws}
\label{appendix:lnk_scaling_law}

\begin{figure}[h]
  \centering
  \includegraphics[width=0.8\textwidth]{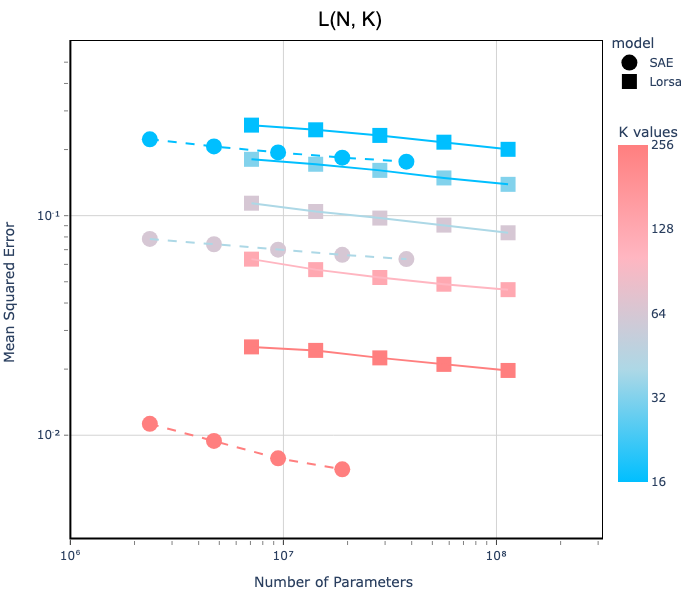}
  \caption{Comparison of scaling law of convergence loss with number of parameters and fixed sparsity(K)
  between SAE and Lorsa trained on layer 3 (out of 12) in Pythia-160M.}
  \label{fig:lnkscalinglaw}
\end{figure}
We explore Low-Rank Sparse Attention scaling laws with respect to both number of learnable parameters $N$ and their
sparsity $K$ (i.e. number of active Lorsa heads per token). Such joint scaling law is similar to TopK Sparse Autoencoders
$L(N, K)$ scaling law reported in~\citet{gao2024oaisae}, which we also replicate for comparison with Lorsa. The $L(N, K)$
scaling laws at layer 3 (out of 12) in Pythia-160M for both SAE and Lorsa are shown in Figure~\ref{fig:lnkscalinglaw}.
It is reasonable that Lorsa has a larger loss than SAE because SAE has identical input and output, making reconstruction easier.
\section{Automated Interpretability Details}
\label{appendix:autointerp}

\paragraph{Evaluation Protocol.}
Our automated interpretability assessment employs a two-phase paradigm adapted from~\citet{bills2023autointerp}:

\begin{enumerate}
    \item \textbf{Explanation Phase}: GPT-4o generates mechanistic explanations using:
    \begin{itemize}
        \item For activation patterns: 8 top-activating token contexts
        \item For $z$-patterns/DFAs: Contribution graphs to max-activating tokens
    \end{itemize}
    
    \item \textbf{Simulation Phase}: GPT-4o predicts activations/patterns for:
    \begin{itemize}
        \item 4 top-activating contexts (testing pattern recognition)
        \item 4 randomly sampled contexts (testing generalization)
    \end{itemize}
\end{enumerate}

\paragraph{Top Activation Explanation Phase Prompt.}

\hfill \break
\begin{promptbox}
  We are analyzing the activation levels of features in a neural network, where each feature activates certain tokens in a text. Each token\'s activation value indicates its relevance to the feature, with higher values showing stronger association. Your task is to infer the common characteristic that these tokens collectively suggest based on their activation values.
  
  Consider the following activations for a feature in the neural network. Activation values are non-negative, with higher values indicating a stronger connection between the token and the feature. Summarize in a single sentence what characteristic the feature is identifying in the text. Don\'t list examples of words. Do not start with "This feature is identifying...". Go straight to the explanation.
  
  Sentence 1:

  <START>

  <|endoftext|><tab>-0.0

  /<tab>-0.0

  */<tab>0.2

  ... (omitted)

  <END>

  Sentence 2:

  ... (omitted)
\end{promptbox}

\paragraph{Top Activation Simulation Phase Prompt.}

\hfill \break

\begin{promptbox}
  We're studying neurons in a neural network. Each neuron looks for certain things in a short document. Your task is to read the explanation of what the neuron does, and predict the neuron's activations for each token in the document.

  For each document, you will see the full text of the document, then the tokens in the document with the activation left blank. You will print the exact same tokens verbatim, but with the activation values filled in according to the explanation. Pay special attention to the explanation's description of the context and order of tokens or words.
  
  Fill out the activation values with integer values from 0 to 10. Don't use negative numbers. Please think carefully. No need to include rationales. Directly start with the first token and do not use code blocks, i.e., ```.
  
  Neuron 1 explanation: This feature is indentifying vowels.
  
  Sequence 1: Tokens without Activations:
  
  a<tab>

  b<tab>

  c<tab>

  d<tab>

  e<tab>

  f<tab>
  
  Sequence 1 Tokens with Activations:
  
  a<tab>10

  b<tab>0

  c<tab>0

  d<tab>0

  e<tab>10

  f<tab>0
  
  Neuron 2 explanation: <Autointerp explanations generated in the previous phase>

  <Few shot examples>
\end{promptbox}

\paragraph{$\vect{z}$ Pattern / DFA Explanation Phase Prompt.}

\hfill \break

\begin{promptbox}
  We are analyzing the attention map of attention heads in a neural network, where each head attends between tokens in a text. Given a head and a query token, we provide each previous token\'s contribution value, with higher values showing stronger association. Your task is to infer the common characteristic of this head that these sequences collectively suggest based on their attention map.
  
  Consider the following attention maps for an attention head. Each line is in the format of <token><tab><value>. Query tokens are additionally highlighted with <token><tab><value><tab>**Query token**. Note that query tokens also attend to themselves. Higher values indicates a stronger contribution from this token to the query token.
  
  Summarize in a single sentence what characteristic the head is attending from and to in the text. It might be helpful to summarize both the commonality of query tokens and source tokens (if any). It is also recommended to mention if this head is often attending to itself.
  
  Don\'t list examples of words. Do not start with "This head is ...". Directly start with the explanation.
  
  Sentence 1: 
  
  <START>
  
  <|endoftext|><tab>-0.0

  /<tab>0.0

  ... (omitted)

  */<tab>0.0<tab>**Query token**
\end{promptbox}

\paragraph{$\vect{z}$ Pattern / DFA Simulation Phase Prompt.}

\hfill \break

\begin{promptbox}
  We're studying attention heads in a neural network. Each head follows a certain attention pattern in a short document. Your task is to read the explanation of what the head does, and predict the head's attention pattern for each previous token in the document, given a specific query token.

  For each document, you will see the full text of the document, then the tokens in the document with the activation left blank. You will print the exact same tokens verbatim, but with the contribution values filled in according to the explanation. Pay special attention to the explanation's description of the context and order of tokens or words.
  
  Each line is in the format of <token><tab>. Query tokens are additionally highlighted with <token><tab>**Query token**<tab>.
  
  Fill out the contribution values with integer values from 0 to 10. Don't use negative numbers. Please think carefully. No need to include rationales. Directly start with the first token and do not use code blocks, i.e., ```.
  
  Head 1 explanation: This head is attending from one vowel to previous vowels and itself.
  
  Sequence 1 Tokens without Activations:
  
  a<tab>

  b<tab>

  c<tab>

  d<tab>

  e<tab>**Query token**
  
  Sequence 1 Tokens with Activations:
  
  a<tab>10

  b<tab>0

  c<tab>0

  d<tab>0

  e<tab>**Query token**<tab>10
  
  Head 2 explanation: <Autointerp explanations generated in the previous phase>

  <Few shot examples>
\end{promptbox}

\section{Additional Case Studies}
\label{appendix:case}

\subsection{Attribution Algorithm for Identifying Lorsa Heads with Specific Functionalities}
\label{appendix:attribution algorithm for identifying lorsa heads with specific functionalities}
In addition to the path patching method discussed in Section~\ref{sec:search_specific_lorsa_heads:rediscover_previously_reported_heads}
, we employ an attribution algorithm, inspired by the approach for detecting important features with attribution 
in~\cite{batson2024attribution}, to identify Lorsa heads associated with specific functionalities. 

The attribution score for a given Lorsa head $h$, is defined as:

$$
  attr_h := O_h \cdot \nabla_x \mathcal{L}
$$

Here, $\nabla_x \mathcal{L}$ is the gradient of the logit on the prediction of the target token with respect to the attention output
$O_h$ of the Lorsa head. For different prompt, we also try logit difference or probability difference to calculate $\nabla_x \mathcal{L}$.

quantifies the contribution of Lorsa head $h$ to the prediction of the correct token.

\subsection{Examples of Lorsa's Rediscovery of Reported Functional Heads}

\label{appendix:examples of lorsa Rediscovered Functional Heads}
\begin{figure}[h]
  \centering
  \includegraphics[width=\textwidth]{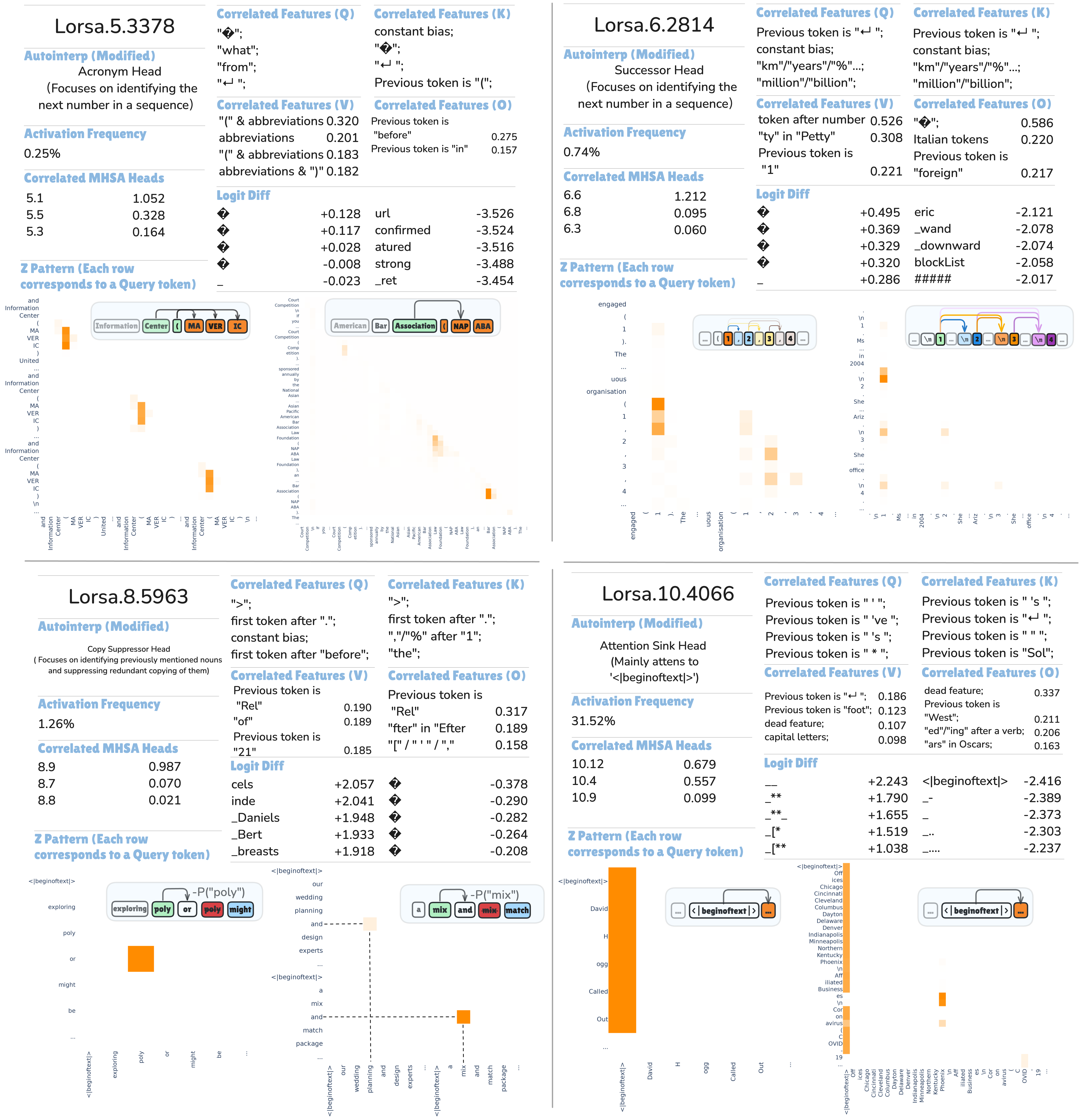}
  \caption{Detailed information on Lorsa's rediscovery of reported functional heads.}
  \label{fig:detailed_lorsa_rediscovered_heads}
\end{figure}
The detailed information on the Lorsa heads discussed in Section~\ref{sec:search_specific_lorsa_heads:rediscover_previously_reported_heads} 
is provided in Figure~\ref{fig:detailed_lorsa_rediscovered_heads}, where we visually demonstrate the logit differences induced by the Lorsa head
,along with the most strongly correlated MSHA heads and SAE features.



\subsection{Arithmetic Lorsa Heads}
\label{appendix:case:arithmetic}

We present the SAE features related to the reported arithmetic Lorsa heads in Table~\ref{tab:arithmetic_head_info},
which shows consistent interpretation in terms of operand plot and $z$ pattern. Additionally, Table~\ref{tab:operator_operand_example_per_id} provides a broader set of examples for these arithmetic Lorsa heads, including functional descriptions and the z-patterns of their top activations.

\begin{table}[h]
  \centering
  \begin{tabular}{@{}ccc@{}}
  \toprule
  Lorsa head ID &
  Manual Interpretation with Operand Plot &
  Manual Interpretation with $z$ Pattern \\ \midrule
  Lorsa.16.20791 & \texttt{op1} $\in \numrange{27}{43}$         & near 30                    \\
  Lorsa.16.20931 & \texttt{op1} $\%\ 10 \in [4, 5, 6]$ & ending with 4 or 6         \\
  Lorsa.16.20947 & \texttt{op1} $\%\ 10 \in [6, 7, 8]$ & ending with 7, sometimes 6 \\
  Lorsa.15.3646  & \texttt{op2} $\%\ 10 =2$            & ending with 2              \\
  Lorsa.15.3813  & \texttt{op2} $\in \numrange{55}{99}$         & from 50 - 99               \\
  Lorsa.15.4001  & \texttt{op2} $\in \numrange{38}{63}$         & near 50                   
  \end{tabular}
  \caption{Supplementary information of Lorsa Head in Figure~\ref{fig:arithmetic_case}.}
  \label{tab:arithmetic_head_info}
  \end{table}
  
  \begin{table}[h]
  \centering
  \begin{tabular}{|c|c|c|c|}
  \hline
  \textbf{ID} & \textbf{Operator} & \textbf{Operand} & \textbf{Top Activation Z Pattern} \\ \hline
  \multirow{4}{*}{Lorsa.15.3646} & Addition & op2 ends with 2 & \multirow{4}{*}{\includegraphics[width=0.25\textwidth]{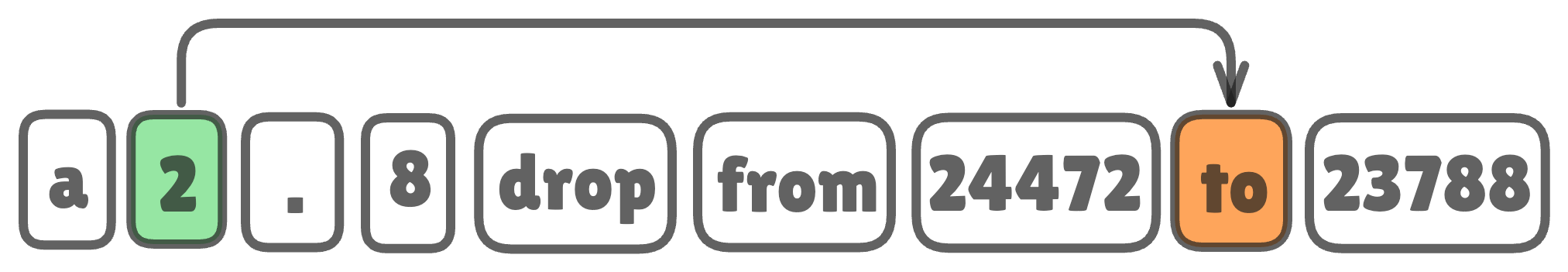}} \\
   & Subtraction & min(op1, op2) ends with 2 &  \\
   & Multiplication & op2 = 2 or 12 &  \\
   & Division & op2 = 2 &  \\ \hline
  \multirow{4}{*}{Lorsa.15.3648} & Addition & op2 ends with 4 & \multirow{4}{*}{\includegraphics[width=0.25\textwidth]{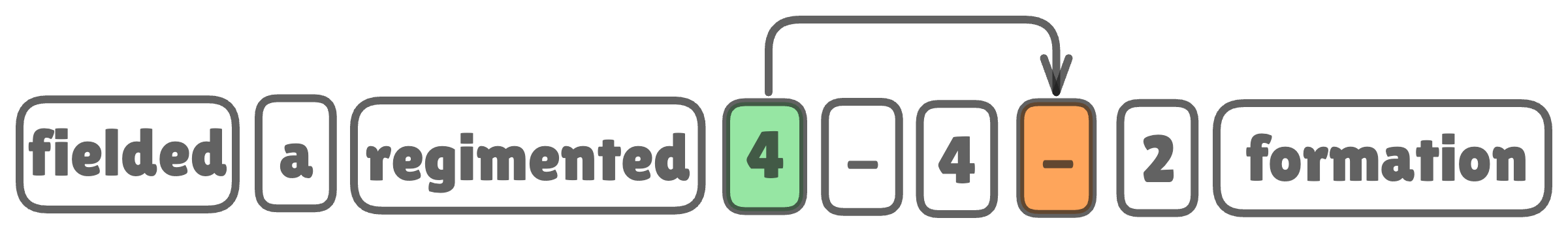}} \\
   & Subtraction & min(op1, op2) ends with 4 &  \\
   & Multiplication & op2 = 4, 24, or 40 &  \\
   & Division & op2 = 4 &  \\ \hline
  \multirow{4}{*}{Lorsa.15.2668} & Addition & Unrelated & \multirow{4}{*}{\includegraphics[width=0.25\textwidth]{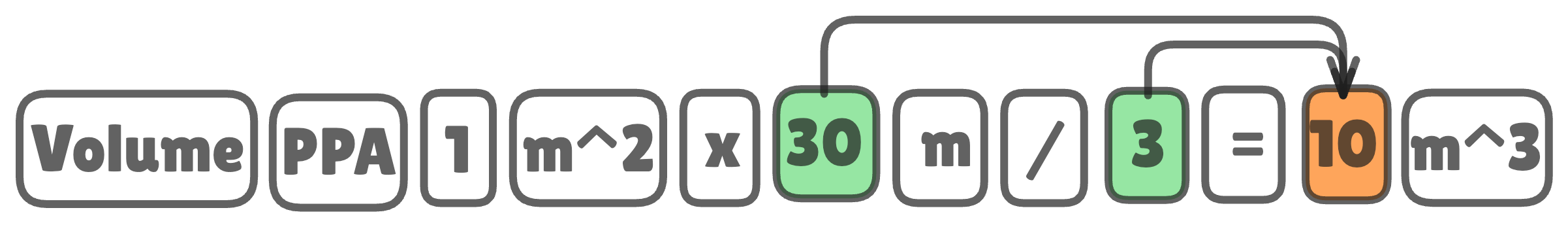}} \\
   & Subtraction & Unrelated &  \\
   & Multiplication & op2 = 3, 6, 30, or 60 &  \\
   & Division & op2 around 3 or 30 &  \\ \hline
  \multirow{4}{*}{Lorsa.15.2770} & Addition & Unrelated & \multirow{4}{*}{\includegraphics[width=0.25\textwidth]{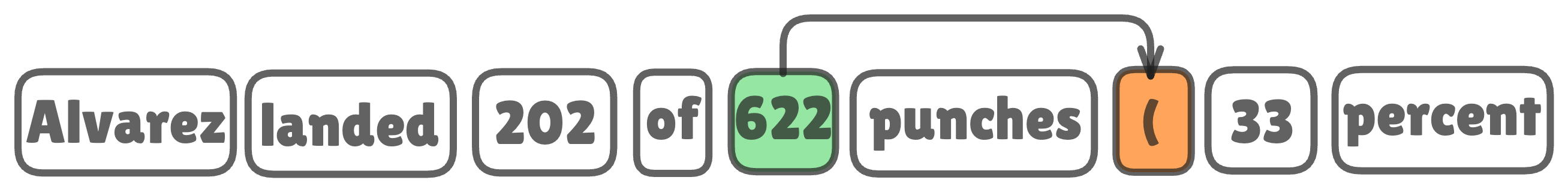}} \\
   & Subtraction & Unrelated &  \\
   & Multiplication & op2 around 62 and its multiples &  \\
   & Division & op2 around 62 and its multiples &  \\ \hline
  \multirow{4}{*}{Lorsa.15.2945} & Addition & Unrelated & \multirow{4}{*}{\includegraphics[width=0.25\textwidth]{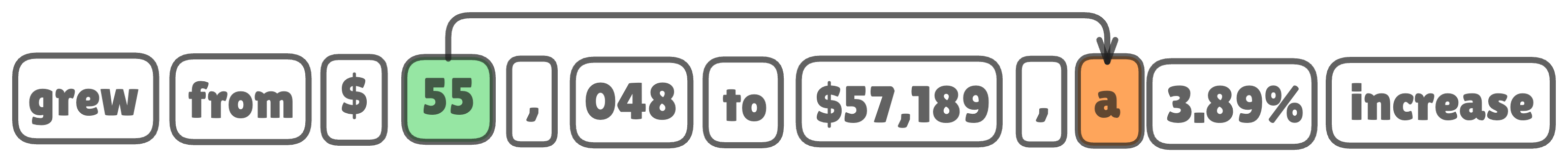}} \\
   & Subtraction & Unrelated &  \\
   & Multiplication & op2 = 7, 11 and their multiples &  \\
   & Division & op2 = 7, 11 and their multiples &  \\ \hline
  \end{tabular}
  \caption{Additional Case of Arithmetic Heads}
  \label{tab:operator_operand_example_per_id}
  \end{table}
  
\section{Assessing Correlation with MHSA}
\label{appendix:corr_wiz_MHSA}

Lorsa is proposed as a method to attack attention superposition. A natural question arises: how is each Lorsa head composed in terms of
the original attention heads? We address this by computing the attribution of each Lorsa head to the original attention heads using an
oblique projection method (Appendix~\ref{appendix:corr_wiz_MHSA:oblique_project}). Analyzing all Lorsa heads trained on Pythia-160M
(Appendix~\ref{appendix:corr_wiz_MHSA:statistics}), we find that roughly half of the Lorsa heads originate from a single original head,
while the other half are superpositions across multiple original heads.

\subsection{Oblique Projection Method for Attribution}
\label{appendix:corr_wiz_MHSA:oblique_project}

Given the output of an original attention head, we project it obliquely onto the (generally non-orthogonal) basis formed by the outputs
of all Lorsa heads at the same layer. The resulting coefficients represent the contribution of the original head to each Lorsa head. Since
the summed outputs of original heads and Lorsa heads closely match, the contribution coefficients for a given Lorsa head approximately sum
to one. Conversely, we similarly compute the fraction of each Lorsa head’s output that can be attributed to each original attention head by
projecting the Lorsa head's output onto the basis formed by the original heads' outputs. All reported results are averaged over more than 1M
tokens.

\subsection{How Many Attention Units are Distributed Across MHSA Heads?}
\label{appendix:corr_wiz_MHSA:statistics}

\begin{figure}[h]
  \centering
  \includegraphics[width=0.7\textwidth]{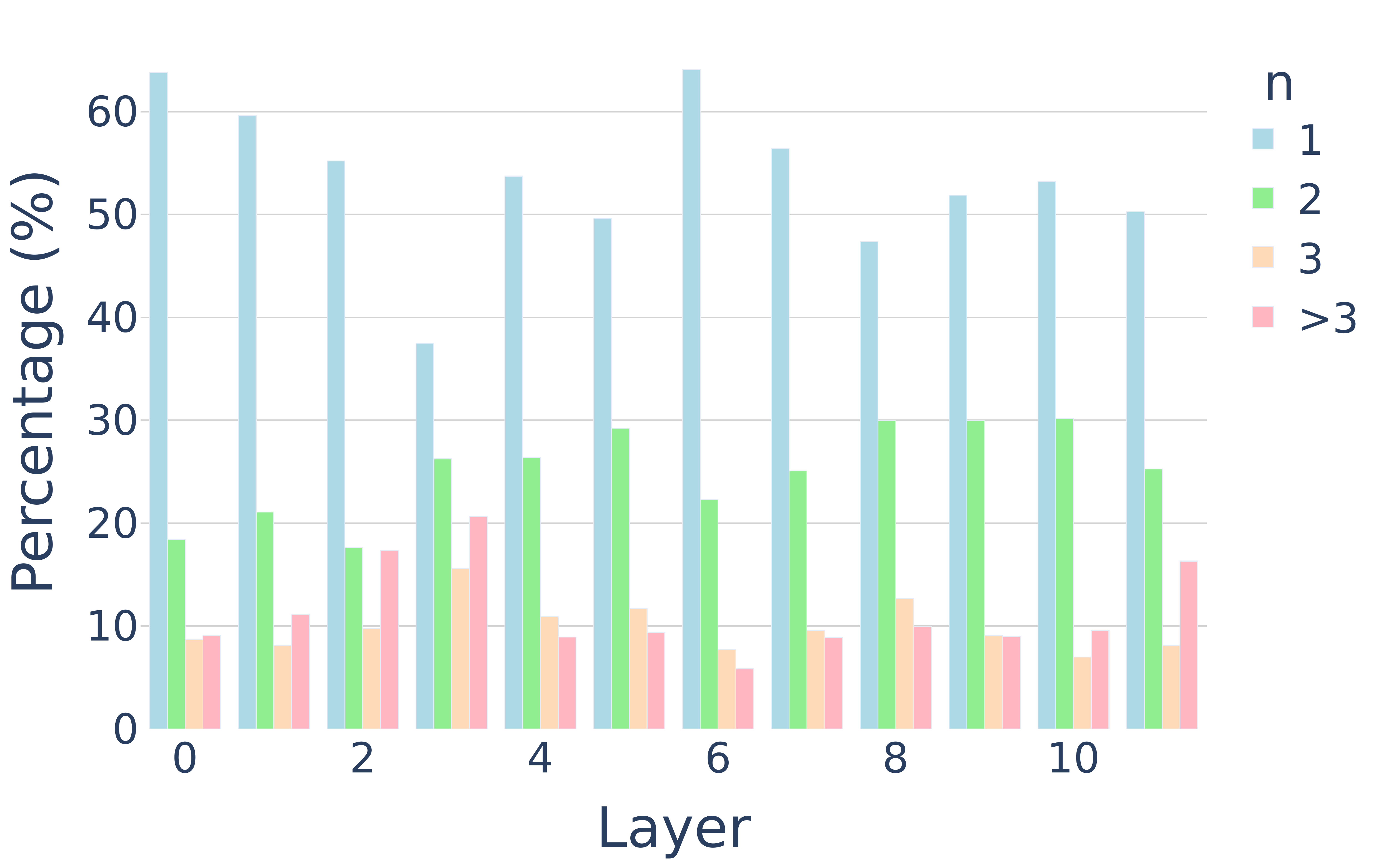}
  \caption{Distribution of Lorsa heads based on the number of original attention heads they are superposed over. No clear trend is observed across different layers. Approximately half of the Lorsa heads are primarily associated with a single original head, about one quarter are superposed over two different original heads, around one eighth are superposed over three different original heads, and the remaining one eighth are superposed over more than three original heads.}
  \label{fig:orig_head_attr_num}
\end{figure}

We compute the attribution statistics for all Lorsa heads trained on Pythia-160M. For a given Lorsa head, we define $n$ as the minimum number
of original heads whose cumulative contributions exceed 90\%. We interpret $n$ as the effective number of original heads a Lorsa head superposes
over. As shown in Figure~\ref{fig:orig_head_attr_num}, approximately half of the Lorsa heads are primarily derived from a single original head,
about a quarter involve two original heads, and the remaining quarter involve three or more original heads.

\subsection{Induction Heads in Pythia-160M}
\label{appendix:induction_heads}

\begin{table}[h]
  \centering
  \caption{Contribution of each MHSA head to induction behavior in Pythia-160M, measured via path patching. Notable induction heads (\texttt{L5.0}, \texttt{L4.6}, \texttt{L5.7}, \texttt{L9.0}, \texttt{L5.6}) are highlighted in bold.}
  \label{tab:induction_heads_pythia160m}
  \resizebox{\textwidth}{!}{
  \begin{tabular}{c|cccccccccccc}
  \toprule
  Layer$\backslash$Head & 0 & 1 & 2 & 3 & 4 & 5 & 6 & 7 & 8 & 9 & 10 & 11 \\
  \midrule
  0  & 0.00 & 0.00 & 0.00 & 0.00 & 0.00 & 0.00 & 0.00 & 0.00 & 0.00 & 0.00 & 0.00 & 0.00 \\
  1  & 0.07 & -0.15 & -0.10 & 0.03 & 0.09 & -0.08 & -0.07 & 0.06 & -0.01 & 0.11 & 0.34 & -0.05 \\
  2  & -0.14 & 0.07 & 0.10 & 0.14 & 0.14 & -0.13 & 0.60 & -0.03 & -0.14 & 0.10 & 0.04 & 0.03 \\
  3  & -0.24 & -0.14 & -0.96 & -1.20 & -0.49 & -0.14 & 0.20 & -0.38 & -0.10 & 0.06 & -0.11 & -0.07 \\
  4  & 0.13 & -0.26 & 0.09 & -0.16 & -0.10 & -0.02 & \textbf{0.89} & 0.13 & 0.09 & -0.28 & -0.14 & 0.30 \\
  5  & \textbf{4.00} & -0.20 & 0.05 & 0.06 & -0.53 & -0.04 & \textbf{0.48} & \textbf{0.62} & 0.06 & 0.08 & 0.05 & -0.23 \\
  6  & -0.04 & -0.23 & -0.04 & -0.22 & 0.02 & 0.09 & 0.04 & -0.33 & 0.02 & -0.04 & -0.38 & 0.04 \\
  7  & -0.28 & 0.17 & 0.03 & 0.06 & -0.28 & -0.07 & 0.01 & -0.18 & -0.23 & -0.03 & -0.02 & 0.18 \\
  8  & -0.07 & 0.03 & 0.50 & 0.00 & 0.15 & -0.02 & 0.01 & -0.22 & 0.02 & -0.02 & -0.08 & 0.38 \\
  9  & \textbf{0.54} & -0.03 & 0.07 & -0.09 & -1.10 & -0.04 & 0.04 & 0.00 & 0.04 & 0.10 & -0.01 & 0.02 \\
  10 & -0.01 & 0.03 & 0.00 & 0.00 & -0.03 & -0.10 & 0.01 & -0.01 & 0.00 & -0.04 & 0.03 & 0.01 \\
  11 & -0.14 & -0.13 & -0.05 & -0.04 & 0.00 & -0.02 & -0.11 & -0.02 & 0.01 & -0.07 & -0.02 & 0.06 \\
  \bottomrule
  \end{tabular}
  }
\end{table}

We use path patching to measure the contribution of each MHSA head in Pythia-160M to induction behavior. The results are shown in Table~\ref{tab:induction_heads_pythia160m}. We find that heads \texttt{L5.0}, \texttt{L4.6}, \texttt{L5.7}, \texttt{L9.0}, \texttt{L5.6} exhibit the most prominent induction signals.

\section{Analyzing SAE Features’ Interaction on Lorsa Heads}
\label{appendix:corr_wiz_SAE_features}
We trained Sparse Autoencoders (SAE) on both the inputs and outputs of Lorsa to facilitate the understanding of its functionality. Since
Lorsa's Q, K, and V are computed from the input, with the output derived from O contributing to the final result, interactions between
SAE features and these components exist across all four aspects: Q, K, O, and V. To evaluate the influence of SAE features on Q and K,
we employ an ablation method (Appendix~\ref{appendix:corr_wiz_SAE_features:q_k}). The correlation between the \(OV \) and SAE features
is assessed using cosine similarity(Appendix~\ref{appendix:corr_wiz_SAE_features:o_v}). For each Lorsa head, we identify the SAE features
most strongly correlated with different aspects. The results are visualized in the Lorsa head dashboard.

\subsection{Quantifying Feature Impacts on Q and K}
\label{appendix:corr_wiz_SAE_features:q_k}
For a given Lorsa head, the impact of a specific feature on Q is calculated as follows: First, we compute the attention pattern at the
activation locations of the Lorsa head. Then, the feature is ablated from the input, and \( Q' \) and the new attention pattern are computed
(with K remaining unaffected). The Kullback-Leibler (KL) divergence between the original and modified attention patterns is used to quantify
the effect of the feature on Q. After iterating over 1 million tokens, the maximum KL divergence observed across all activations of the Lorsa
head is taken as the measure of the feature's influence on Q for this head. A similar approach is used to calculate the impact of a feature
on K, with the difference being that when recalculating the attention pattern, all instances of K are recomputed using the modified input,
while Q remains unchanged.

\subsection{Quantifying Feature Correlations with O and V}
\label{appendix:corr_wiz_SAE_features:o_v}
For a given Lorsa head, both the weight vectors \( W_O \) and \( W_V \) are one-dimensional vectors of size \( D_{model} \). Therefore, for
each SAE feature trained on the Lorsa input, the contribution to \( V \) is linear, meaning that the contribution of each feature to \( V \)
scales proportionally with the feature's activation value. Similarly, for each activation \( z \) of the head, the contribution of SAE features
trained on the Lorsa output to the activation value is also linear. We compute the cosine similarity between the decoder of each SAE feature
trained on the Lorsa input and \( W_V \), which quantifies its correlation with \( V \) for the given Lorsa head. Similarly, the cosine
similarity between the encoder of each SAE feature trained on the Lorsa output and \( W_O \) is computed to measure its correlation with
\( O \) for the given Lorsa head.

\section{Lorsa Dark Matter}
\label{appendix:dark_matter}
\begin{figure}[htbp]
  \centering
  \begin{subfigure}[b]{0.8\textwidth}
    \centering
    \includegraphics[width=\linewidth]{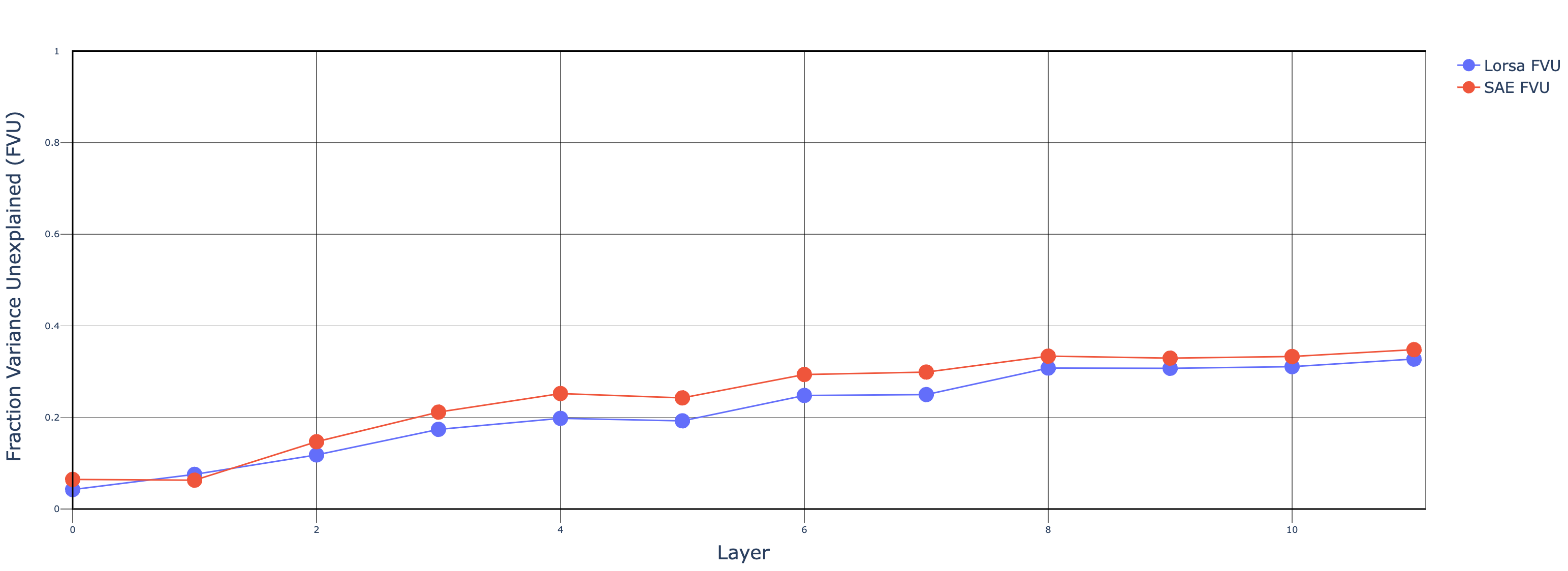}
    \caption{FVU for Each Layer in Pythia-160M}
    \label{fig:subfig1}
  \end{subfigure}


  \begin{subfigure}[b]{0.8\textwidth}
    \centering
    \includegraphics[width=\linewidth]{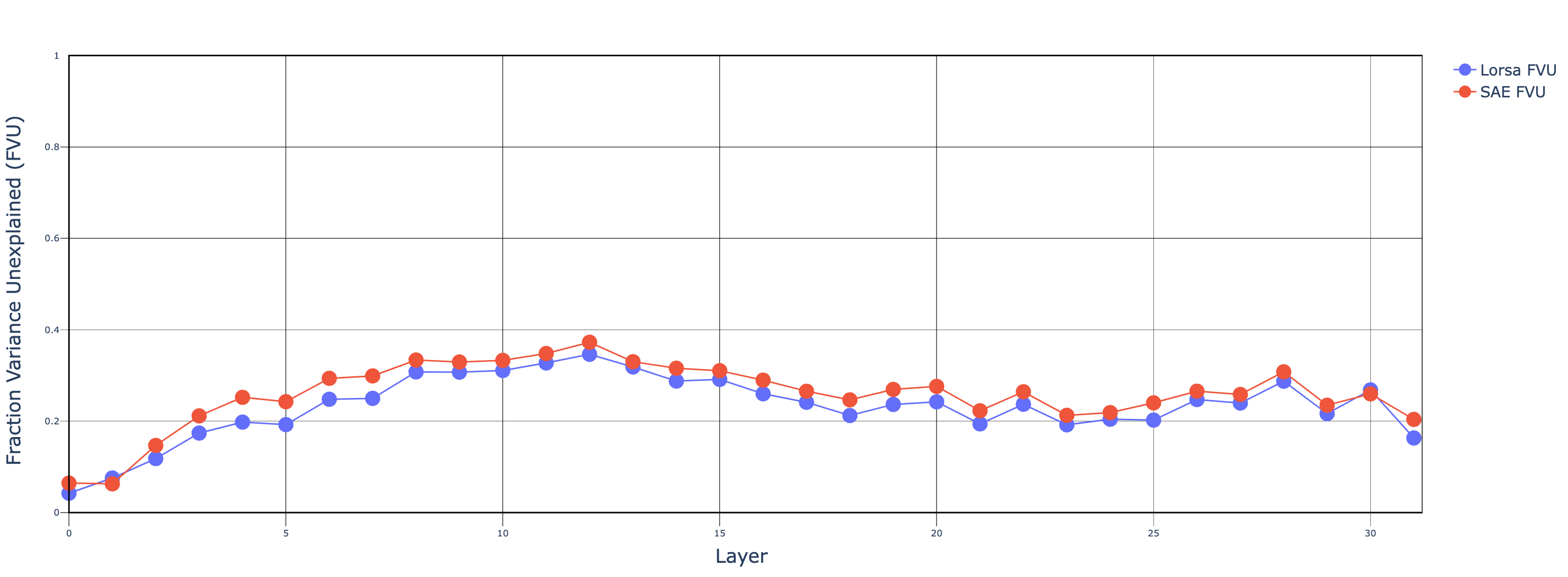}
    \caption{FVU for Each Layer in Llama-3.1-8B}
    \label{fig:subfig2}
  \end{subfigure}

  \caption{FVU of Lorsa and SAE across each layer in Pythia-160M and Llama-3.1-8B. Both models show a similar trend in FVU.}
  \label{fig:fvu_pythia_llama}
\end{figure}

\begin{figure}[h!]
  \centering
  \includegraphics[width=0.8\textwidth]{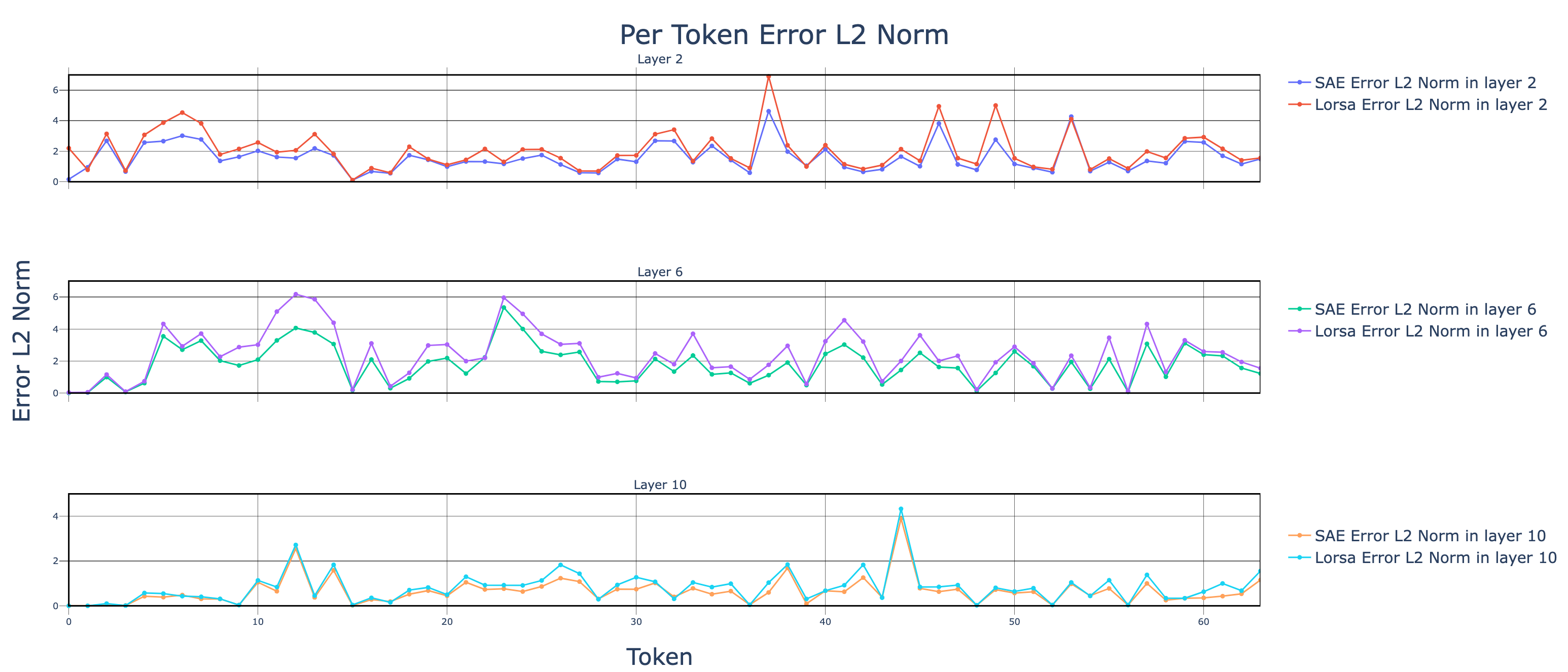}
  \caption{Per-Token Error Norms of Lorsa and SAE on Layers 2, 6, and 10 of Pythia-160M for 64 tokens.}
  \label{fig:per_token_loss}
\end{figure}

\begin{figure}[t!]  
  \centering
  \vspace*{-\baselineskip}  
  \includegraphics[width=0.8\textwidth]{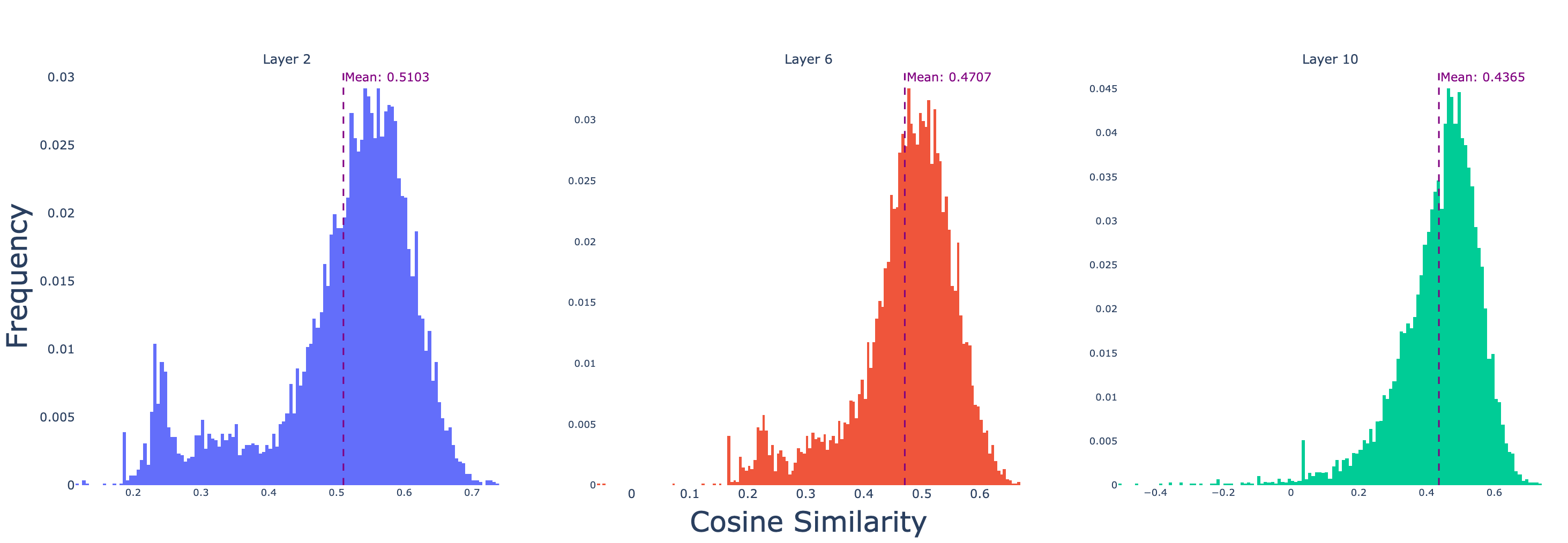}
  \caption{Per-Token error cosine similarity distribution between Lorsa and SAE on Layers 2, 6, and 10 of Pythia-160M, measured on approximately 10,000 Tokens.}
  \label{fig:cos_similarity}
\end{figure}

In Figure~\ref{fig:fvu_pythia_llama}, we examine the Fraction of Variance Unexplained (FVU) for both Lorsa and SAE on the same attention layers
of Pythia-160M and Llama-3.1-8B. 
To further explore error patterns, Figure~\ref{fig:per_token_loss} illustrates the per-token error norms of Lorsa and SAE across layers 2, 6,
and 10 of Pythia-160M on a set of 64 tokens. 
Figure~\ref{fig:cos_similarity} then quantifies the distribution of cosine similarity between Lorsa and SAE's per-token error norms on the
same layers, measured on approximately 10,000 tokens. These results indicate that the loss pattern between pre token between Lorsa and SAE
has a nontrivial correlation. 

It is interesting that both Lorsa and SAE exhibit a positive correlation in their magnitudes and trends for FVU and per-token error norms.

We propose that this is not a coincidence, and hypothesize that it stems from a shared gap between sparse dictionary learning and the representation
structure of data within the model. 
Alternatively, this correlation may arise from the challenge that sparse dictionary learning faces in capturing super-rare data features or certain
nonlinear or dense components within the features.

While we observe that Lorsa generally yields a higher FVU and error norm compared to SAE, this could suggest that Lorsa captures a greater amount of
"dark matter" relative to SAE. This distinction arises because SAE is trained to reconstruct activations, while Lorsa is optimized to predict the
actions of the original attention heads.

\section{Towards Full Sparsification of A 2-Layer Transformer}
\label{appendix:fully_sparsify}

\begin{figure}[h!]
  \centering
  \includegraphics[width=0.5\textwidth]{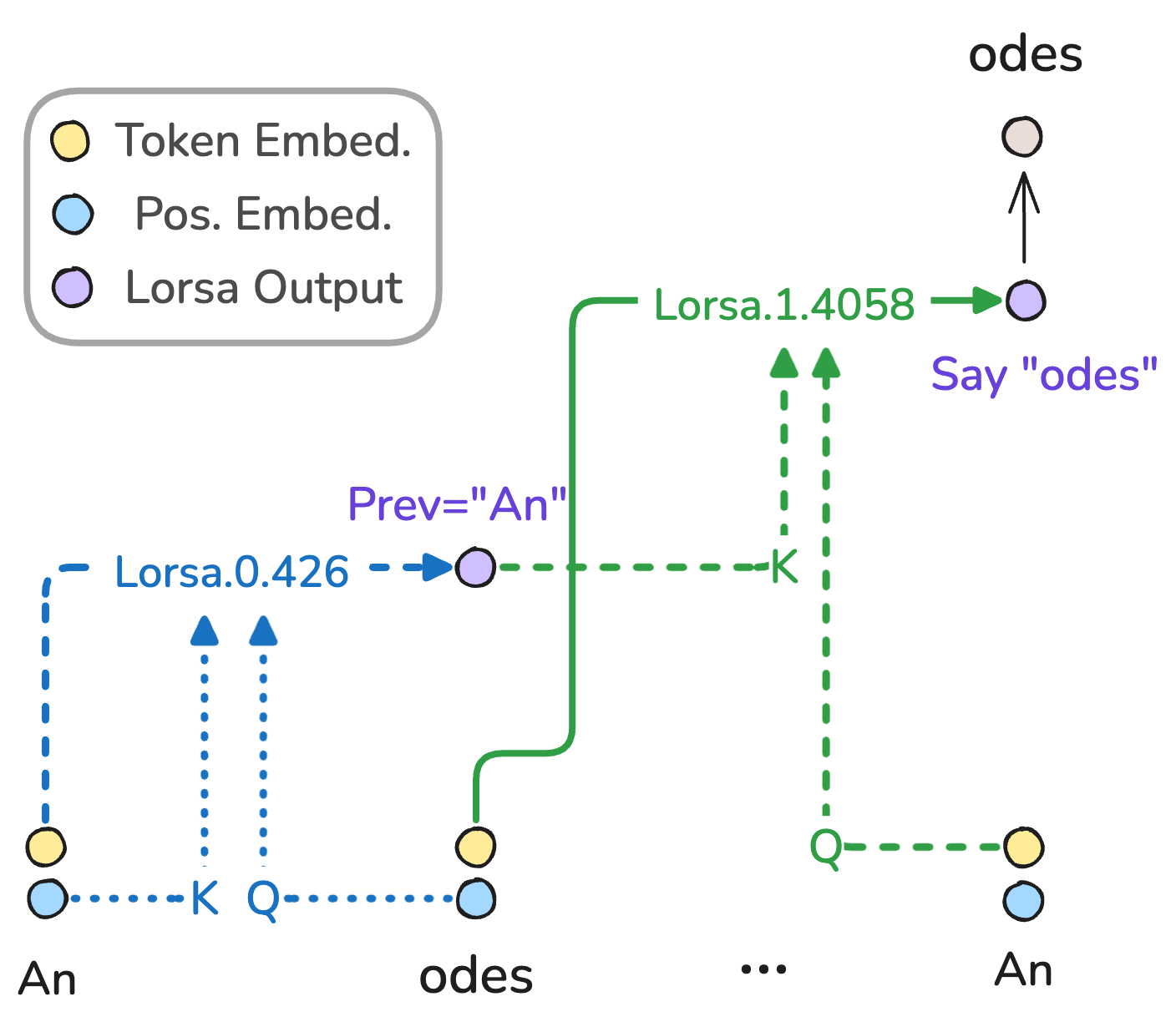}
  \caption{Induction circuit found in our fully sparsified replacement model.}
  \label{fig:full_sparsify}
\end{figure}

Since our final goal is to understand Transformers' inner working by breaking down MHSA and MLPs into atomic units (Figure~\ref{fig:head_img}),
we train Lorsa and Transcoder~\citep{dunefsky2024transcoder} on a 2-layer Transfomer
(\href{https://huggingface.co/NeelNanda/GELU_2L512W_C4_Code}{link}). We follow the method introduced in~\citet{ge2024hierattr}
where they multiply features via QK circuit to find the most salient feature pairs contributing to QK scores. Alternatively applying
attribution through Transcoder features / Lorsa heads and QK ablation gives us the clear attribution graph for induction behavior
(Figure~\ref{fig:full_sparsify}). Due to the capability constraint of this model, we failed
to observe more interesting behaviors or attribution graphs involving Transcoder features.
Nonetheless, we believe applying Lorsa and Cross-Layer Transcoders~\citep{ameisen2025circuit} to a larger model may reveal a
lot of surprising behaviors, following the spirit of~\citet{lindsey2025biology}.

\end{document}